\documentclass{article} % For LaTeX2e
\usepackage[preprint]{neurips_2026}
\usepackage{times}
\usepackage[most]{tcolorbox}
\usepackage[hidelinks]{hyperref}

\usepackage{amsmath,amsfonts,bm}

\def\eqref#1{equation~\ref{#1}}
\def\1{\bm{1}}

\DeclareMathAlphabet{\mathsfit}{\encodingdefault}{\sfdefault}{m}{sl}
\SetMathAlphabet{\mathsfit}{bold}{\encodingdefault}{\sfdefault}{bx}{n}

\newcommand{\R}{\mathbb{R}}

\usepackage{hyperref}
\usepackage{url}

\title{Zero-shot generalization of transformer neural operators to larger domains}

\author{%
Armand de Villeroché$^{1}$ \quad Sibo Cheng$^{1}$ \quad Vincent Le Guen$^{2,3}$ \quad Marc Bocquet$^{1}$\\
\textbf{Rem-Sophia Mouradi}$^{3}$ \quad \textbf{Patrick Armand}$^{4}$ \quad \textbf{Alban Farchi}$^{1}$ \quad \textbf{Patrick Massin}$^{1}$\\
$^{1}$CEREA, ENPC, EDF R\&D, Institut Polytechnique de Paris, Île-de-France, France\\
$^{2}$SINCLAIR AI Laboratory, Saclay, Île-de-France, France\\
$^{3}$EDF R\&D, Île-de-France, France\\
$^{4}$CEA, DAM, DIF, F-91297 Arpajon, France\\
\texttt{armand.de-villeroche@edf.fr}
}

\usepackage{booktabs}       % professional-quality tables
\usepackage{siunitx}
\newcommand{\G}[0]{\mathcal{G}}

\newcommand{\concat}[0]{\bigoplus}

\newcommand{\xmark}{%
\tikz[scale=0.23] {
    \draw[color=red!70!black,line width=0.7,line cap=round]
        (0,0) to [bend left=6] (1,1);
    \draw[color=red!70!black,line width=0.7,line cap=round]
        (0.2,0.95) to [bend right=3] (0.8,0.05);
}}
\newcommand{\cmark}{%
\tikz[scale=0.23] {
    \draw[color=green!60!black,line width=0.7,line cap=round]
        (0.25,0) to [bend left=10] (1,1);
    \draw[color=green!60!black,line width=0.8,line cap=round]
        (0,0.35) to [bend right=1] (0.23,0);
}}

\begin{document}

\definecolor{mydarkblue}{rgb}{0, 0.08, 0.45}
\hypersetup{
    colorlinks=true,
    citecolor=mydarkblue,
    linkcolor=mydarkblue,
    urlcolor=mydarkblue
}

\definecolor{acadblue}{RGB}{235, 240, 250}
\tcbset{
    blue_style/.style={
        colback=acadblue,       
        colframe=black,         
        boxrule=0.5pt,          
        arc=1pt,                
        auto outer arc,
        left=10pt, right=10pt, top=5pt, bottom=5pt, 
        enhanced,
        before skip=10pt,       
        after skip=10pt         
    }
}

%\iclrfinalcopy % Uncomment for camera-ready version, but NOT for submission.

\maketitle

\begin{abstract}

Transformer-based neural operators have shown remarkable performance for approximating solution operators of partial differential equations on complex geometries. However, existing approaches implicitly assume a fixed domain size, which limits their ability to generalize at inference. In this work, we investigate domain extension, namely zero-shot inference on spatial domains that are significantly larger than those encountered during training. We argue that this setting fundamentally requires spatial locality and translation equivariance. We propose to implement this locality via a decomposable bias in the attention logits computation, enabling finely controllable locality while remaining fully decomposable into query–key inner products and directly compatible with optimized attention kernels. Combined with rotary positional embeddings, it enables expressive embeddings with controllable spatial support without altering the transformer architecture. We empirically show that our approach substantially improves zero-shot generalization to larger domains across two PDE benchmarks and a 3D industrial atmospheric flow application. Our code and data is available at https://github.com/cerea-daml/domain-extension.

\end{abstract}

\section{Introduction}

Partial differential equations (PDE) are ubiquitous in physics and engineering, but are often computationally intensive to solve with traditional numerical solvers. Deep learning based \textit{neural operators} \citep{Kovachki2023} have emerged as an attractive alternative: once trained, they offer fast inference while capturing complex nonlinear behaviors in high‑dimensional systems. Yet this perspective hinges on large amounts of high-fidelity training data, which can be prohibitively expensive to obtain. As the computational cost of a simulation scales with the domain size, training a model on simulations over small domains while retaining accuracy over larger domains is a promising alternative.

Among existing neural operator architectures, transformer‑based models \citep{Cao2021} have shown strong performance across a variety of benchmarks \citep{Alkin2024, Wu2024, Villeroche2026, Wen2025}. By applying attention directly to unstructured point clouds, transformers provide flexibility for complex geometries and can scale to real-world industrial problems \citep{Alkin2025}. Despite these advances, current neural operator architectures are only trained and evaluated on domains of predefined size, which limits their applicability at inference time.

\paragraph{Domain extension}

We consider the problem of \textit{domain extension}, defined as zero-shot generalization to a substantially larger domain than the training domain. This problem is particularly challenging: while tasks such as super‑resolution or inference from partial observations correspond to an interpolation problem, domain extension is fundamentally an extrapolation problem. Despite its practical importance, this setting is barely explored for neural operators with prior work limited to domain decomposition \citep{Huang2025} or graph neural network models \citep{Pfaff2020}.

\paragraph{Locality vs globality}
To be universal, a neural operator must be non-local \citep{Lanthaler2025, Calvello2025}, i.e. point-wise predictions must depend on data over the entire domain. Yet in practice neural operators are trained on domains of finite size, and hence can only learn dependencies \textit{up to that size}. For transformer-based neural operators, non-locality is achieved via an integration operator, whose underlying integration kernel support will be \textit{implicitly bounded} by the finite training domain. However, this implicit assumption breaks if the size of the domain changes at inference time. Hence, we propose to make this support bounded \textit{explicitly}. This makes the operator's locality to the size of the training domain explicit and controllable, rather than implicit, and brings robustness to changes of the domain shape at inference, such as increasing the domain size.

\paragraph{Locality is hard to implement in practice}

Non-periodic physical fields have long been represented using localized representations such as wavelets \citep{Daubechies1992}, which offer a principled way to model spatially bounded interactions. However, integrating true wavelet‑like or compactly supported locality into attention mechanisms remains difficult: numerically efficient attention kernels \citep{Dao2022} assume that attention logits decompose into simple query–key inner products. Any positional structure that violates this constraint typically requires custom GPU kernels, which are challenging to generalize to multi‑dimensional unstructured point clouds and become a serious barrier to scalability in large‑scale neural operator applications.

% Alternatively, one can rely on distance-based attention biases, which can be written as inner products of the query and key positions provided the correct distance function is chosen \citep{Wu2025}. Yet, such biases are less expressive than more commonly used Rotary Positional Embeddings (RoPE) \citep{Su2024}. 
This tension between the need for spatial locality, expressiveness and the constraints imposed by efficient attention implementations motivates the search for positional encodings that are simultaneously relative, localized, expressive, and fully compatible with optimized computational kernels. Ideally, such encodings should enable a tunable spatial support while preserving the efficiency of high‑throughput matrix‑multiplication‑based attention.

To address this challenge, we combine Rotary Positional Encoding (RoPE) \citep{Su2024} with a decomposable distance bias \citep{Wu2025} (Figure \ref{fig:intro}). This creates a highly expressive relative position embedding with a tunable localized support, while ensuring compatibility with existing efficient attention kernels. We further extend existing positional biases using hyperbolic functions, creating numerically stable, asymmetric biases while retaining decomposability properties. Our main contributions are:

\begin{figure}
    \centering
    \includegraphics[clip, trim=0.15cm 4.85cm 0.35cm 4.25cm, width=\linewidth]{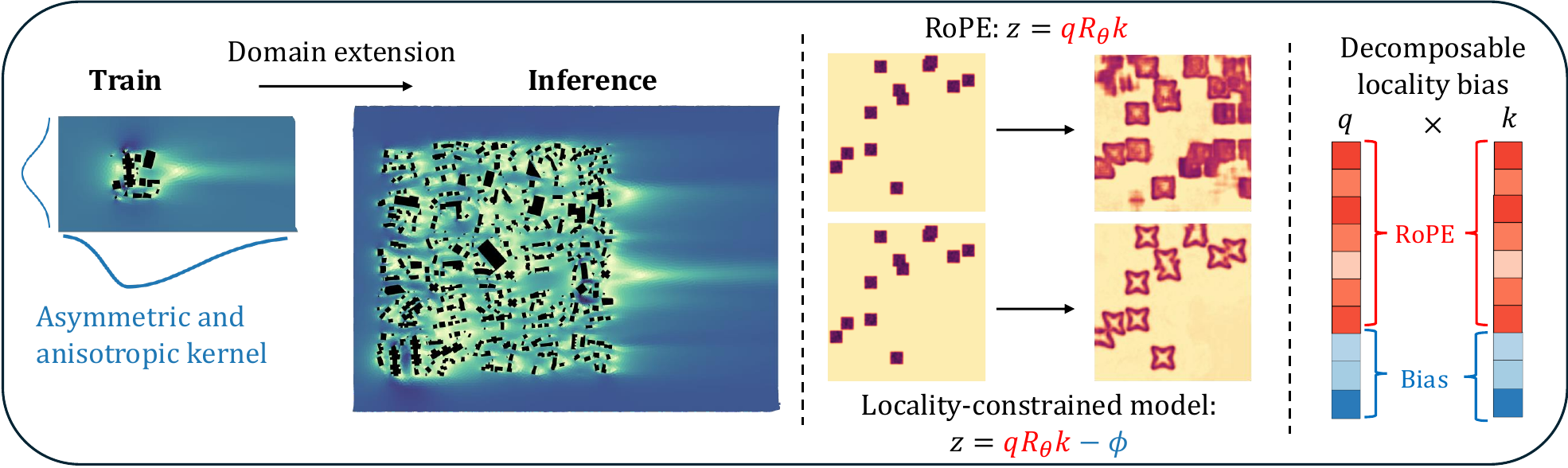}
    \caption{Domain extension. When increasing the domain size, RoPE creates spurious interactions for distances larger than the training domain. Our approach removes these interactions by enforcing locality via a decomposable bias term, enabling zero-shot generalization to larger domains without modification to the attention mechanism. This constraint must be fitted anisotropically and asymmetrically to the training domain to preserve long-range interactions.}
    \label{fig:intro}
\end{figure}

\begin{itemize}
    \item we formulate the problem of \textit{domain extension}, which is critical for scaling up to industrial-level applications, and show that it can be solved assuming a \textit{local} neural operator;
    \item we combine RoPE with decomposable distance biases to create expressive position embeddings with localized kernel support that remains compatible with numerically efficient attention implementations. We propose formulations for both symmetric kernel support and asymmetric kernel support;
    % \item we show that an L2 distance bias restricts the kernel support with a Gaussian function, and propose an extension that allows to restrict the support both anisotropically and asymmetrically;
    \item we demonstrate empirically that this approach enables inference on substantially larger domains than training across two 1D and 2D benchmark problems and a challenging 3D industrial use case with a complex geometry.
\end{itemize}

\section{Locality via decomposable positional biases} \label{sec:method}

\subsection{Preliminary}

\paragraph{Notations}

Let $\Omega \subset \R^p$ be a bounded domain. We adopt a neural operator view, and consider a continuous \textit{latent field} $u:\Omega \to \R^d$, with $\R^d$ the latent space. 
We write $q,k,v:\Omega \to \R^d$ the query, key and value fields, obtained by learnable point-wise linear projections from the latent field $u$, and $z:\R^d\times\R^d\times\Omega\times\Omega\to\R$ the function used to compute the attention logits between two points. We note that the expression of $z$ will depend on the positional embedding used. 
Finally, we adopt the variables $\xi\in\Omega$ to write the position of a key point and $c\in\Omega$ to write the position of a query point.
%\textcolor{red}{You should introduce the position vectors c and $\xi$ of $\Omega$ here.} 

\paragraph{Attention as an integral operator}

Standard single-head scaled dot-product attention \citep{Vaswani2017} is a Monte-Carlo approximation of the integral operator $\mathcal{G}$ defined as \citep{Cao2021, Kovachki2023}: 
\begin{equation}
    \label{eq:int-operator}
    \forall c \in \Omega, \,\mathcal{G}(u)(c)
    = \int_\Omega \kappa\!\left(z(q(c),k(\xi),c,\xi)\right) v(\xi)\, \mathrm{d}\xi,
\end{equation}
where the kernel $\kappa$ is defined by
\begin{equation}
    \label{eq:kappa}
    \kappa(z(q(c),k(\xi),c,\xi)) = \frac{\exp(z(q(c),k(\xi),c,\xi) / \sqrt{d})}
     {\int_\Omega \exp(z(q(c),k(\xi'),c,\xi') / \sqrt{d}) \, \mathrm{d}\xi'}.
\end{equation}

\subsection{Requirements}

\paragraph{Decomposability}

Numerical computation of the kernel $\kappa$ implies a quadratic memory cost with the number of points over which $u$ has been discretized, which is prohibitive for a large sequence of points. To reduce the memory footprint, optimized computation kernels, such as FlashAttention \citep{Dao2022}, decompose the computation using tiled matrix multiplications. While bringing considerable memory overhead reduction and computational speedup, fused GPU implementations of these algorithms have limited support for logits computation operations that are not written as a scalar product of a query and a key vector, making such an operation a barrier for scalability. Hence, to avoid this limitation, the logits function $z$ must be \textit{decomposable} as the scalar product of a function of $(q(c),c)$ and a function of $(k(\xi),\xi)$.

\paragraph{Locality and translation invariance}

We aim at preserving the translation invariance of $\kappa$ — i.e., $\kappa = \kappa(\cdot, \cdot, c - \xi)$ depends only on relative positional differences. This property is particularly important for zero-shot generalization in large-scale physical domains. We therefore construct the new attention mechanism on top of RoPE as the backbone.

Nevertheless, RoPE and its variants rely on periodic functions to encode relative distances. When a model is trained on a bounded domain $\Omega_{\textrm{train}}$ and scaled up in a zero-shot manner to a significantly larger domain $\Omega_{\textrm{test}}$, the periodic nature of the embedding can induce ambiguities in distance representation \citep{Press2021, Peng2023}. As a result, large relative distances in $\Omega_{\textrm{test}}$ may be mapped to similar embeddings as for shorter distances observed during training, thereby artificially inducing attention between points at distances that were not encountered within $\Omega_{\textrm{train}}$.

Instead, we seek to restrict the support of $\kappa$ to only interactions that were seen during the training phase. We argue that this can be archived by a \textit{local} model. We can define the ensemble of oriented distances seen during training as $S_\mathrm{train} = \{c-\xi \mid (c,\xi)\in\Omega_\mathrm{train}^2\}$. We then require the kernel's support to be local, and bounded by $S_\mathrm{train}$: $\kappa(\cdot, \cdot, c - \xi) = 0$ whenever $c - \xi \notin S_\mathrm{train}$. As a consequence,
\begin{tcolorbox}[blue_style]
\[
\G(u)(c)= \int_{\Omega_\mathrm{test}} \kappa(z(q(c),k(\xi),c-\xi))v(\xi)\mathrm{d}\xi = \int_{S(c)} \kappa(z(q(c),k(\xi),c-\xi))v(\xi)\mathrm{d}\xi,
\]
\end{tcolorbox}
where $S(c)=\{\xi \mid c-\xi \in S_\mathrm{train}\}$. Thus, at a given position $c$, $\mathcal{G}(u)(c)$ depends only on information within its neighborhood $S(c)$ and is therefore independent of the test domain $\Omega_{\mathrm{test}}$. Thus $\mathcal{G}$ remains robust under enlargements of $\Omega_{\mathrm{test}}$ relative to $\Omega_{\mathrm{train}}$, without requiring retraining. Intuitively, this approach can be understood as \textit{preserving the same kernel} when changing the domain size at inference time.

% To enforce the locality of the integration operator, we impose a compact support on $\kappa$. In this paper, we define the locality of the attention mechanism as follows: there exists $r \in \mathbb{R}^+$ such that, for all $c,\xi \in (\Omega_{\mathrm{train}} \cup \Omega_{\mathrm{test}})^2$, $\kappa(\cdot, \cdot, c - \xi) = 0$ whenever $|c - \xi| > r$. As a consequence,
% \begin{tcolorbox}[blue_style]
% \[
% \G(u)(c)= \int_{\Omega_\mathrm{test}} \kappa(z(q(c),k(\xi),c-\xi))v(\xi)\mathrm{d}\xi = \int_{\B_r(c)} \kappa(z(q(c),k(\xi),c-\xi))v(\xi)\mathrm{d}\xi,
% \]
% \end{tcolorbox}
% where $\B_r(c) = \{\xi \mid |c - \xi| < r\}$.
% Thus, at a given position $c$, $\mathcal{G}(u)(c)$ depends only on information within its neighborhood $\B_r(c)$ and is therefore independent of the test domain $\Omega_{\mathrm{test}}$. Thus $\mathcal{G}$ remains robust under enlargements of $\Omega_{\mathrm{test}}$ relative to $\Omega_{\mathrm{train}}$, without requiring retraining.

\paragraph{Shape of the kernel support}

% We can express the ensemble of all possible oriented distances present in the training domain as $S = \{c-\xi \mid (c,\xi)\in\Omega_\mathrm{train}^2\}$. 
Ideally, the kernel $\kappa$ should have a support that closely matches $S_\mathrm{train}$, such that it is sufficiently large to capture all interactions present in $\Omega_\mathrm{train}$ during training, but avoids introducing spurious long-range interactions during zero-shot domain extension. To achieve this, the support of the kernel should be able to match closely $S_\mathrm{train}$. We highlight here two desirable properties of $\kappa$:

- \textit{Anisotropy}: as $\Omega_\mathrm{train}$ may be non-isotropic in complex geometries, $S_\mathrm{train}$ may reflect this structure. Consequently, $\kappa$ should be anisotropically restrictable. 

- \textit{Asymmetry}: asymmetry may arise when $\Omega_\mathrm{train}$ is not homogeneous and contains a sub-area $\Omega_\mathrm{train}^b \subset \Omega_\mathrm{train}$ which presents a specific behavior that determines the final solution. A common example is fluid-solid interactions: obstacles impacting the flow are often placed upwind in a larger simulated domain. This situation can also arise from the neural operator structure when cross-attention or perceiver attention is used to couple different parts of the simulation domain, as is done in \citep{Alkin2025, Villeroche2026}. 
Under this assumption, the solution's behavior will be determined by values of the key function within $\Omega_\mathrm{train}^b$: consequently, relevant interactions will be dominated by $S^b_\mathrm{train}=\{c-\xi \mid c\in\Omega_\mathrm{train}, \xi\in\Omega_\mathrm{train}^b\}$. Unlike the generic homogeneous case, $S^b_\mathrm{train}$ is \textit{not necessarily centered around $0$}. To represent this, $\kappa$ may need a different, asymmetric support that accounts for the relative position of $c$ and $\xi$. 

\subsection{Locality via a position bias} \label{sec:local-pe}

Commonly used RoPE \citep{Su2024} embeds positions via a rotation matrix $R_\theta \in \R^{d \times d}$, formally defined in Appendix \ref{app:rope}. This formulation corresponds to a Fourier basis decomposition of the logits function on frequency basis $\theta$ with modulated coefficients constructed from $q(c)$ and $k(\xi)$ (see Appendix \ref{app:rope}). While particularly expressive and decomposable, this approach is inherently non-local.

\begin{table}[h]
    \centering
    \caption{Different bias functions. $m \in \R^+, r \in \R^+$ and $A_+, A_-$ diagonal matrices with positive coefficients are hyperparameters associated with each method, $\alpha$ being a token-wise learnable weight, and $\{e_i\}$ an  orthonormal basis of $\Omega$. *FlashBias in its original formulation is isotropic; in this work, we extend it to an anisotropic formulation.}
    \label{tab:methods}
    \small
\centering
    \begin{tabular}{llccc} \toprule
        &&& \multicolumn{2}{c}{Kernel support} \\ \cmidrule{4-5}
        Method & Logits bias
        $(z = q^\top R_\theta^\top k + \sqrt d \textcolor{blue}{b})$ & Decomp. & Aniso. & Asym. \\ \midrule
        % RoPE & $b=0$ & \xmark & \cmark  & -\\
        Window attention & $ b =  - \infty \1 (\|c-\xi\|_1>r)$ & \xmark & \xmark & \xmark \\
        ALiBi & $b = - m \|c-\xi\|_1$ & \xmark & \xmark & \xmark \\
        FlashBias & $b = - \alpha(u(c))\|c-\xi\|_2^2$ & \cmark & \cmark* & \xmark \\
        % LASPE (ours) & $b =  - \frac{\sqrt d}{2} (c-\xi)^\top A (c-\xi) $ & \cmark & \cmark & \xmark \\
        Hyperbolic bias (ours) & $ b = - \frac{1}{2} \sum_i e^{A_-^{1/2} (c-\xi) \cdot e_i} + e^{-A_+^{1/2} (c-\xi)\cdot e_i} $ & \cmark & \cmark & \cmark \\
        \bottomrule
    \end{tabular}
\end{table}

% To create a localized embedding that preserves the high expressiveness of RoPE, we add locality via a distance based position bias added to the logits. % Furthermore, for the method to be applicable with optimized attention kernels, we require this bias to be decomposable as well.
% Table \ref{tab:methods} subsumes commonly used position biases in transformers. Window attention \citep{Beltagy2020} and ALiBi \citep{Press2021} respectively apply a sliding attention window of half-width $r\in\R^+$ or a linear bias with distance with coefficient $\alpha\in\R^+$. While these methods could be applied to enforce locality, they are not decomposable and hence would require significant implementation and computation overhead. We instead focus on decomposable bias functions: FlashBias \citep{Wu2025} propose to use the L2 distance to create a decomposable distance bias. With hyperparameter $A = \mathrm{diag}(\lambda_1^{-2}, \dots \lambda_p^{-2}) \in \R^{p \times p}$ a diagonal matrix with positive coefficients, we extend this bias to anisotropic configurations and define Local Anisotropic Symmetric Positional Encoding (LASPE) as:

To create a localized embedding that preserves the high expressiveness of RoPE, we add locality via a distance based position bias added to the logits. Unlike previous bias-based position embeddings \citep{Press2021, Wu2025} which rely on a bias as the main position embedding methods, we view the bias purely as a geometrical term controlling locality, and retain RoPE as the position embedding backbone.

Table \ref{tab:methods} summarizes commonly used position biases in transformers. Window attention \citep{Beltagy2020} and ALiBi \citep{Press2021} respectively apply a sliding attention window of half-width $r\in\R^+$ or a linear bias with distance with coefficient $m\in\R^+$. While these methods could be applied to enforce locality, they are not decomposable and hence would require significant implementation and computation overhead. We instead focus on decomposable bias functions: FlashBias \citep{Wu2025} propose to use the L2 distance to create a decomposable distance bias. We extend this bias to anisotropic configurations and define the associated logits as:
\begin{equation}
    \label{eq:pos-emb-sym}
    z^\mathrm{L2} = q^\top(c)R^\top_\theta(c-\xi)k(\xi) - \frac{\sqrt d}{2} (c-\xi)^\top A(c-\xi),
\end{equation}
where the hyperparameter $A = \mathrm{diag}(\lambda_1^{-2}, \dots \lambda_p^{-2}) \in \R^{p \times p}$ is a diagonal matrix with positive coefficients controlling anisotropy. For $A= \mathrm{Id}$, the bias term corresponds to FlashBias with a fixed coefficient $\alpha=1/2$.
As this expression relies on a symmetric distance function $(c-\xi)^\top A(c-\xi)$, it cannot be used to define an asymmetric bias. We further propose a new formulation based on the hyperbolic function $\cosh$:
\begin{tcolorbox}[blue_style]
\begin{equation}
    \label{eq:pos-emb-asym}
    z^\mathrm{hyp} = q^\top(c)R^\top_\theta(c-\xi)k(\xi) - \frac{\sqrt d}{2} \sum_i \left\{ \exp(A_-^{1/2} (c-\xi) \cdot e_i)+\exp(A_+^{1/2} (\xi-c) \cdot e_i) \right\},
\end{equation}
\end{tcolorbox}
where $A_+=\mathrm{diag}(\lambda_{1,+}^{-2}, \dots \lambda_{p,+}^{-2})\in\R^{p\times p}$ and $A_-=\mathrm{diag}(\lambda_{1,-}^{-2}, \dots \lambda_{p,-}^{-2})\in\R^{p\times p}$ are two different anisotropy matrices inducing \textit{asymmetric} behaviors depending on the relative position of $c$ and $\xi$, and $\{e_i\}$ can be any orthonormal basis of $\R^p$. We show in Appendix \ref{app:local-pe} that both methods can be decomposed into query-key inner products.

In the symmetric case $A_+=A_-=A$, our formulation closely matches the original symmetric formulation of an L2-distance bias: indeed we show in Appendix \ref{app:local-pe} that $z^\mathrm{L2} \sim z^\mathrm{hyp}$ up to an additive constant if $(c-\xi)\cdot e_i \ll \lambda_i$. For $(c-\xi)\cdot e_i \gg \lambda_i$, both decay to $-\infty$. Consequently, we consider our formulation to be a natural extension of the L2 symmetric bias for asymmetric cases.

\paragraph{Choosing the kernel width}

Selecting the width of the kernel correctly is critical in our method: a too small kernel will not model long distance interactions present in the training domain, while a too large kernel will be sensitive to an increase in domain size. An ideal kernel' support should match closely the ensemble of interactions present in the training domain $S_\mathrm{train}$. However, a position bias does not enforce a strict limit of the kernel support, but instead imposes exponential decay to $0$. To better build an intuition, we consider the $1D$, symmetric and isotropic case with a coefficient $\lambda \in \R^+$. Then the integration kernel reads:
\[
f_\lambda(c-\xi) = e^{-\frac{(c-\xi)^2}{2\lambda^2}}, 
\qquad 
\kappa(z(q(c), k(\xi), c, \xi)) = \frac{ f_\lambda(c-\xi) e^{q^\top(c)R_\theta^\top(c-\xi)k(\xi)/\sqrt d}}{\int_\Omega  f_\lambda(c-\xi')e^{q^\top(c)R_\theta^\top(c-\xi')k(\xi') / \sqrt d}\mathrm{d}\xi'}.
\]

Hence, the integration kernel support is controlled via a \textit{Gaussian function of standard deviation of $\lambda$}. Under this insight, we consider the attention weights to become statistically insignificant for a distance exceeding $3\lambda$. Consequently, \textit{$\lambda$ should be chosen at most as a third of the training domain size} to restrict the integration kernel to interactions present in the training data.

The above expression can be easily extended to multiple dimensions, for which $f$ will be a multivariate Gaussian function with separate standard deviations per axis determined by the coefficients of $A$.

% We also note that window attention could be decomposably approximated using $L_P$ norm bias with $P \gg 2$, which results in $P$-order gaussians, which in turn approximates the crenel form of an attention window. However, this would require using $P(P+1)/2$ feature coefficients to generate the bias term, inducing a high computation overhead. Hence this approach is not considered in this work. %Should we keep this paragraph? It's a cozy observation, but space is scarse. %TODO

\section{Numerical experiments}
\label{sec:experiments}

We validate our approach on 1D and 2D academic datasets and a 3D industrial use case. For each dataset, we evaluate the zero-shot robustness of the trained model on progressively larger test domains.

\subsection{Academic datasets}

As currently existing benchmarks do not have validation simulations over large domains, we generate our own datasets on classically-evaluated PDEs: Shallow Water equations and Gray-Scott equations. Shallow water equations model the evolution of the water height $h$ and the velocity $v$ of a water surface in 1D. Gray-Scott equations model the evolution of concentrations $U$ and $V$ in a reaction-diffusion system in 2D. For each dataset, we generate a training set of simulations over domain $\Omega_\mathrm{train}=[0,1000]^p$ in normalized units, with $p$ the physical system dimension. We then generate large scale test datasets, with a per-dimension scaling of the domain size controlled by a factor $s_p$: $\Omega_\mathrm{test}^s=[0,1000s_p]^p$. Further information on the solved PDE and datasets can be found in Appendix \ref{app:datasets}, and visualization of the simulations are available in Appendix \ref{app:visu}.

\begin{figure}[t]
    \centering
    \includegraphics[width=\linewidth]{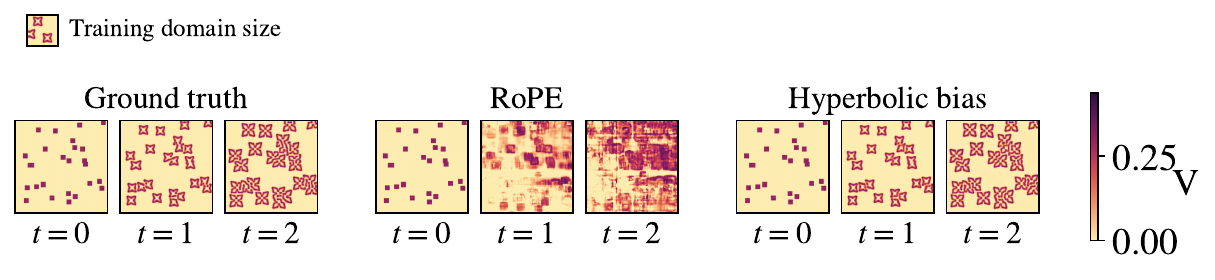}
    \caption{Autoregressive prediction of $V$ for 3 steps of the Gray-Scott PDE for $s_x \times s_y =3 \times 3$.}
    \label{fig:gray-scott}
\end{figure}

To show the universality of our approach, we use a generic transformer neural operator architecture consisting of stacked transformer blocks with skip connections, linear input embeddings, and linear output projections. The model and hyperparameters are further described in Appendix \ref{app:training}. We do not use absolute position embeddings on the initial embeddings, and only embed the previous time step and a boundary indicator. Combined with relative attention positional embeddings, this makes the model equivariant by translation, and robust to new coordinates when changing the domain size. 
We compare models trained with RoPE, an L2 bias, and a hyperbolic bias in Table \ref{tab:academic-results}. For the L2 and the hyperbolic biases, we use an isotropic decay length of $\lambda = 250$, yielding a kernel support comparable to the training domain size. For both datasets, RoPE performance degrades significantly as the inference domain size increases, while errors for locality‑constrained models remain nearly constant.
The L2 bias shows increased error at higher scalings ($s \ge 7$), which we attribute to numerical instabilities and further discuss in Appendix \ref{app:num-stability}. The relatively high base error for Gray–Scott is due to small average values of $\Delta V$, which amplify the normalized $L_1$ metric. Figure \ref{fig:gray-scott} shows that an hyperbolic bias accurately predicts $V$ autoregressively over $3$ time steps for $s_x \times s_y = 3 \times 3$, whereas unbiased RoPE predictions degrade immediately.

\begin{table}[h]
    \centering
    \caption{L1 error [\%] for different test domain size where $s$ denotes the scaling factor of the test domain size compared to the one of training. $\Delta\cdot$ denotes the difference between two time steps.}
    \label{tab:academic-results}
    \begin{tabular}{c*{5}{cc}} \toprule
         \multicolumn{11}{c}{\textbf{Shallow water equations} (1D+time)} \\
         $s_x$ & \multicolumn{2}{c}{1} & \multicolumn{2}{c}{2} & \multicolumn{2}{c}{3}  & \multicolumn{2}{c}{7} & \multicolumn{2}{c}{10} \\ \cmidrule(lr){2-3} \cmidrule(lr){4-5} \cmidrule(lr){6-7} \cmidrule(lr){8-9} \cmidrule(lr){10-11}
         & $\Delta v$ & $\Delta h$ & $\Delta v$ & $\Delta h$ & $\Delta v$ & $\Delta h$ & $\Delta v$ & $\Delta h$ & $\Delta v$ & $\Delta h$ \\
         RoPE &  1.1 & 1.2  &  1.4 & 1.5  &  2.2 & 2.3  &   7.4 & 7.4  &  11.4 & 10.8 \\
         L2 bias &  1.4 & 1.4  &  1.3 & 1.3  &  1.3 & 1.3 &  2.2 & 2.2  &  4.5 & 4.4 \\
         Hyperbolic bias &  1.2 & 1.2  &  1.2 & 1.2  &  1.2 & 1.2 &  1.1 & 1.1  &  1.1 & 1.1 \\

         \midrule
         \multicolumn{11}{c}{\textbf{Gray-Scott equations} (2D+time)} \\
         $s_x \times s_y$ & \multicolumn{2}{c}{$1\times1$} & \multicolumn{2}{c}{$2\times2$} & \multicolumn{2}{c}{$3\times3$}  & \multicolumn{2}{c}{$7\times7$} & \multicolumn{2}{c}{$10\times10$} \\ \cmidrule(lr){2-3} \cmidrule(lr){4-5} \cmidrule(lr){6-7} \cmidrule(lr){8-9} \cmidrule(lr){10-11}
         & $\Delta U$ & $\Delta V$ & $\Delta U$ & $\Delta V$ & $\Delta U$ & $\Delta V$ & $\Delta U$ & $\Delta V$ & $\Delta U$ & $\Delta V$ \\
         RoPE &  31 & 33  &  153 & 175  &  347 & 373  &  436 & 483  &  386 & 430   \\
         L2 bias &  28 & 30  &  25 & 28  &  26 & 29  &  67 & 78  &  119 & 136 \\
         Hyperbolic bias &  29 & 32  &  26 & 30  &  26 & 30  &  27 & 30  &  27 & 31   \\
         
         \bottomrule         
    \end{tabular}
\end{table}

\subsection{Microscale atmospheric flow}

We further evaluate our approach on a challenging industrial use case such as microscale atmospheric flow in urban environments. We rely on the AB‑SWIFT model \citep{Villeroche2026}, a specialized model for atmospheric flow built upon AB‑UPT \citep{Alkin2025}, a well‑validated recent architecture for PDE learning. We use the associated RandomBuildingsDataset, which consists of 3D steady‑state simulations of wind flow around randomly generated urban geometries. Beyond increased geometric and physical complexity, this task exhibits strong anisotropy due to the rectangular domain and a pronounced asymmetry, as buildings are located on the upwind side of the simulation domain. As a result, learnable interaction lengths differ substantially upstream and downstream of the building area. Moreover, predicting steady states involves long‑range correlations, making this task a stringent test of locality‑constrained operators.

\begin{figure}[t]
    \centering
    \includegraphics[clip, trim=0.0cm 6.8cm 0.0cm 4.25cm, width=\linewidth]{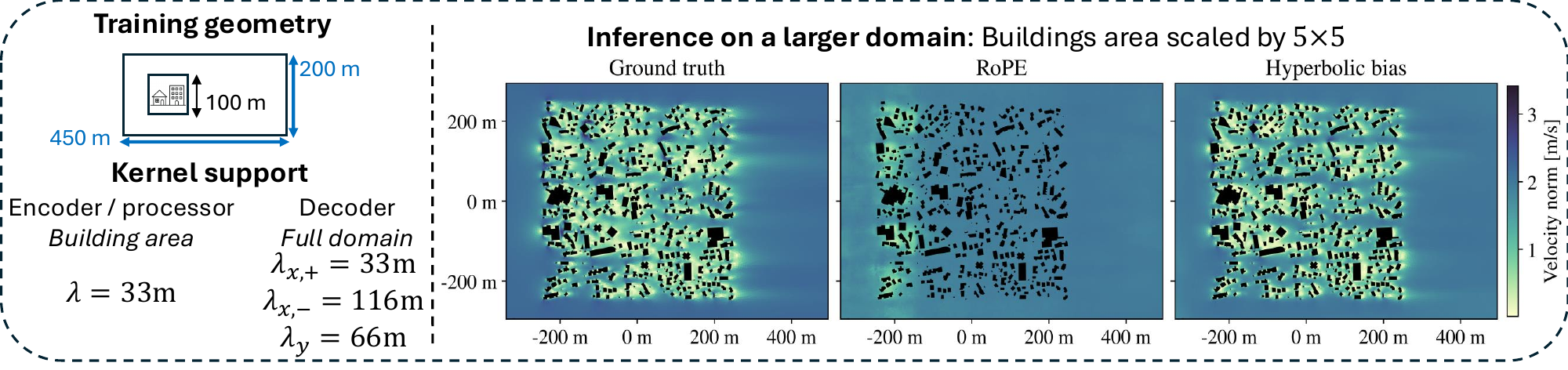}
    \caption{Left: geometry of the industrial use case with buildings located in an off-centered, squared area upwind of the domain. To reflect this, the geometry encoder and the processor are parametrized to have a kernel support limited to the size of the buildings area, and the volume decoder has an off-centered asymmetric kernel corresponding to the size of the full volumic mesh. Right: prediction of AB-SWIFT with an hyperbolic bias or unbiased RoPE, with a scaling factor of $s_x \times s_y=5\times 5$.}
    \label{fig:abswift}
\end{figure}

AB‑SWIFT follows a multi‑block architecture with encoder, processor, and decoder modules, each operating over different spatial regions. The encoder embeds geometric information localized to the building area, the processor jointly handles geometric and volumetric information over the full domain, and the decoder operates solely on volumetric data. To reflect the heterogeneous geometry and interaction patterns of each module, we use distinct kernel supports per block. We use hyperbolic biases to allow for asymmetries, and to avoid numerical stability issues associated with L2 biases (Appendix \ref{app:num-stability}). The encoder and processor employ an isotropic and symmetric support matched to the size of the square building region. In contrast the decoder uses an asymmetric and anisotropic support reflecting the shape of the domain and the upstream/downstream asymmetry. %In contrast the decoder acts on the rectangular wake region, with an upstream/downstream asymmetry. 

% To assess robustness under domain extension, we generate new validation cases following the CFD setup of \citet{Villeroche2026}, scaling the building region by factors of $s_x \times s_y = 2 \times2$ and $s_x \times s_y = 5 \times5$, and generating $10$ new simulations per scale. 
% We restrict our study to neutrally stratified atmospheric conditions, as other stratifications yield boundary conditions that interact strongly with the buildings due to their proximity to the upstream domain boundary, and which could not be reproduced reliably when increasing the domain size.

We compare AB‑SWIFT models trained with RoPE or an hyperbolic bias on test domains with scalings of $s_x \times s_y = 2 \times 2 $ and $ 5 \times 5$ of the buildings sub-area. Data generation is described in Appendix \ref{app:datasets}. To preserve translation equivariance, the model omits absolute positional embeddings; all other hyperparameters are identical to the original AB-SWIFT implementation. Performance is evaluated using the normalized L1 error on the velocity field, reported separately for the building and wake regions due to their scale‑dependent relative sizes. Table \ref{tab:abswift-results} shows that the error increases with domain size when using an hyperbolic bias, albeit much less than with RoPE. Furthermore, Figure \ref{fig:abswift} shows that the locality-constrained model preserve the overall flow structure, yielding a coherent prediction, while RoPE prediction fails catastrophically. Finally, the local model still captures the long-range wakes behind the buildings area, showing that it is still capable of modeling long-range correlations, up to the size of the kernel support. Consequently, we attribute the error increase for the local model metrics to new flow patterns arising from having buildings in a larger area, and which are not present in the original training data.

\begin{table}[h]
    \caption{L1 error [\%] on the velocity field for different domain scaling factors. The building area is the square region occupied by buildings, while the wake area comprises the rest of the domain.}
    \label{tab:abswift-results}
    \centering
    \begin{tabular}{ccccccc} \toprule
         \multicolumn{7}{c}{\textbf{Microscale atmospheric flow}} \\
         & \multicolumn{3}{c}{Buildings area} & \multicolumn{3}{c}{Wakes area} \\ \cmidrule(lr){2-4} \cmidrule(lr){5-7}
         $s_x\times s_y$ & $1\times1$ & $2\times2$ & $5\times5$ & $1\times1$ & $2\times2$ & $5\times5$ \\
         RoPE & 4.9 & 13.8 & 33.0   &   1.3 & 5.5 & 8.2 \\
         Hyperbolic bias & 4.5 & 9.4 & 16.7   &   1.2 & 3.8 & 5.3 \\ %L1 vel error with averaging with repeat wrapper of 10.
         \bottomrule
    \end{tabular}
\end{table}

\section{Limitations} \label{sec:limitations}

\paragraph{Sparsity} An additional advantage of the proposed framework is that locality induces sparsity in the attention matrix, which has the potential to significantly reduce computational cost, ultimately enabling linear complexity in the number of points \citep{Beltagy2020}. However, in this proof-of-concept work, we do not yet exploit this structure, and our implementation retains the quadratic complexity of full self-attention. However, while windowed attention in natural language processing (NLP) yields an easily exploitable banded matrix structure, unstructured, multi-dimensional point clouds typically lead to randomly sparse attention matrices. This makes sparsity significantly harder to exploit in practice. An interesting direction for future work could be combining our work with algorithms reordering the mesh to enforce a banded structure such as space-filling curve, Cuthill-McKee mesh reordering \citep{Cuthill1969} or ball tree partitioning, which have already been used for attention mechanisms \citep{Zhdanov2025}.

\paragraph{Correlations longer than the training domain} While we show that locality makes neural operators independent of the inference domain size, we note that the true solution operator may not be: the underlying physics may present correlations at a longer range than what is seen in the training domain, which will not be captured by our approach. However, to capture these correlations in a zero-shot generalization setting, the learned integration kernel would have to be extrapolated and generalized beyond its training support. While we note that some physics may well-behave enough to make this extrapolation possible, it is not true in general, as fundamental solution behaviors may vary depending on the scale of the problem. A more robust approach is to fine-tune a model on a few large cases, which we leave to future work.

\section{Related work}

We review related work through the lens of \emph{domain extension} for neural operators. While many studies address generalization across resolutions, geometries, or partial observations, domain extension constitutes a fundamental extrapolation setting that has received comparatively little attention.

\paragraph{Extension in NLP via representation scaling}
A related problem arises in natural language processing, where transformers struggle to extrapolate beyond the context lengths seen during training. Methods such as ALiBi~\citep{Press2021} introduce distance‑dependent biases into attention logits, while more recent approaches rely on position interpolation~\citep{Chen2023,Peng2023} to rescale positional coordinates for longer contexts. Despite conceptual similarities, these approaches assume scale‑invariant, one‑dimensional inputs—an assumption that breaks for PDEs defined over multi‑dimensional domains with physically meaningful interaction lengths. 

\paragraph{Architectural scaling via domain decomposition}
Domain decomposition methods~\citep{Toselli2004,Mathew2008} scale numerical solvers by partitioning large domains into smaller subdomains, and have recently been adapted to neural operators~\citep{Huang2025}. However, these approaches operate at the application level rather than at the operator level, requiring explicit coupling mechanisms and introducing challenges when stitching subdomain interfaces. They do not directly address the behavior of a single learned operator under domain enlargement. In contrast, our work targets robustness to domain extension without requiring explicit partitioning or retraining.

% \paragraph{Physics-informed global operators}
% Neural operators \citep{Kovachki2023}  learn mappings between function spaces for fast PDE surrogates. Fourier Neural Operators \citep{Li2023} use global spectral convolutions to model long‑range interactions but are restricted to regular grids and fixed domain sizes. Variants such as PINO \citep{Li2024} or FC‑PINO \citep{Ganeshram2022} improve physical consistency and robustness via PDE constraints and frequency‑domain regularization, yet retain fully global formulations and lack explicit control over spatial interaction ranges, limiting zero‑shot extrapolation to larger domains.

\paragraph{Locality in other neural operator architectures} Other neural operator architectures rely on localized convolution kernel. Graph neural operators \citep{Li2020a} are natively local, while \citet{Liu-Schiaffini2024} add localized kernel to Fourier Neural operators \citep{Li2023}. However these approaches are still limited by the shortcomings of their specific architectures; graph neural networks struggle to computationally scale to large meshes or model long-distance effects while fourier neural operators lack support for unstructured meshes.

\paragraph{Kernel-level control through locality and positional encodings}
Transformer‑based neural operators~\citep{Cao2021,Alkin2024,Alkin2025,Wu2024, Serrano2024} operate directly on unstructured point clouds and provide highly expressive non‑local representations. Relative positional encodings such as RoPE~\citep{Su2024,Heo2024} are widely adopted due to their translation equivariance and compatibility with optimized attention implementations such as FlashAttention~\citep{Dao2022}. However, their quasi-periodic structure can associate large relative distances with shorter ones, inducing spurious interactions under domain extension. Locality can be enforced explicitly through window attention~\citep{Beltagy2020}, but strong attention masks break decomposability and limit scalability on irregular geometries. More recently, FlashBias~\citep{Wu2025} introduces a decomposable L2 distance bias compatible with efficient attention kernels, providing controlled locality but restricting the kernel support to symmetric configurations.

Our work builds on these developments by introducing explicit, decomposable control over the spatial support of attention integration kernels. By extending distance‑based biases to anisotropic and asymmetric configurations, we enforce physically meaningful locality while preserving expressiveness and compatibility with high‑performance attention kernels, enabling robust zero‑shot domain extension.

\section{Conclusion and future work}

In this study, we addressed the problem of domain extension for transformer-based neural operators by introducing explicit spatial locality into attention mechanisms. We have shown that an L2 bias restricts kernel support to a Gaussian-like shape, and proposed a novel hyperbolic decomposable bias that extends existing methods to asymmetric settings, while avoiding numerical stability issues of an L2 bias decomposition. Experiments on multiple PDE benchmarks, including a large-scale 3D industrial case, show that enforcing locality significantly improves zero-shot generalization to larger domains.

More broadly, our results suggest that enforcing structured locality in neural operators may provide a principled route toward appropriate scaling laws in machine learning. By explicitly constraining the effective support of interactions, models can generalize to larger domains without requiring proportional increases in training data or model capacity. This points to an alternative scaling paradigm where performance gains arise not only from data and model scale, but from modulating inductive biases that reflect how information propagates across space.

%%%%%%%%%%%%%%%%%%%%%%%  Ethics and such %%%%%%%%%%%%%%%%%%%%%%%%%%%%%%
\subsection*{AI use statement}

% (This section is \textbf{required} and does not count toward the page limit.)

In this work we have only used generative AI for corrective editing and proofreading of the paper's text, and for creating the plotting scripts for some of the figures presented in this paper. All AI-generated texts and edits have been thoroughly reviewed and verified. We take responsibility for the final content of this work, including text, claims or artifacts produced with the aid of generative AI.

\subsection*{Reproducibility statement}

% (This section is \textbf{recommended} and does not count toward the page limit.)

% It is important that the work published in ICLR is reproducible. Authors are
% strongly encouraged to include a paragraph-long Reproducibility Statement at the
% end of the main text (before references) to discuss the efforts that have been
% made to ensure reproducibility. This paragraph should not itself describe
% details needed for reproducing the results, but rather reference the parts of
% the main paper, appendix, and supplemental materials that will help with
% reproducibility. For example, for novel models or algorithms, a link to an
% anonymous downloadable source code can be submitted as supplementary materials;
% for theoretical results, clear explanations of any assumptions and a complete
% proof of the claims can be included in the appendix; for any datasets used in
% the experiments, a complete description of the data processing steps can be
% provided in the supplementary materials. Each of the above are examples of
% things that can be referenced in the reproducibility statement.

Section \ref{sec:local-pe} describes the methodology used in this paper, while Annex \ref{app:local-pe} describes the associated implementation. Appendix \ref{app:training} describes the training and models setup, and Appendix \ref{app:datasets} the dataset generation processes. Furthermore, all codes and datasets are available at https://github.com/cerea-daml/domain-extension.


\subsubsection*{Acknowledgments}
This work was funded by the French National Association of Research and Technology (ANRT), EDF R\&D and the CEA DASE with the Industrial Conventions for Training through REsearch (CIFRE grant agreement 2023/1614). The authors acknowledge their support, as well as SINCLAIR AI lab for helpful discussions.

{ 
% \small
% \bibliographystyle{iclr2027_conference}
\bibliographystyle{apalike}
\bibliography{pe_paper}
}

%%%%%%%%%%%%%%%%%%%%%%%%%%%% Appendix %%%%%%%%%%%%%%%%%%%%%%%%%%%%%%%%

\appendix
\section{Position embeddings}

This appendix presents detailed formulations and decomposed expression of all position embeddings used in this work.

\paragraph{Setting}

Let $p$ denote the spatial dimension of the physical domain (typically 2 or 3). We consider a canonical spatial basis $\{e_1,\dots,e_p\}$. Let $c, \xi \in \R^p$ be two spatial locations, with $c_i=c\cdot e_i$ and $\xi_i = \xi \cdot  e_i$. Let $q(c), k(\xi) \in \R^d$ be the query and key embeddings of a given attention head.

We denote $\concat$ the concatenation operation of two or more tensors.

\subsection{Rotary positional embedding} \label{app:rope}

Rotary positional embeddings (RoPE) were first introduced by \citet{Su2024} in the context of natural language processing, and later generalized to multi‑dimensional settings by \citet{Heo2024}. RoPE embeds positional information using rotation matrices. Let $\theta \in \R^{d/2 \times p}$ denote a (possibly learnable) basis of frequencies. We define the block‑wise rotation matrix $R_\theta$ as
\[
    R_\theta(c) =
    \begin{pmatrix}
        r(\theta_0 \cdot c) & & & 0 \\
        & r(\theta_1 \cdot c) & & \\
        & & \ddots & \\
        0 & & & r(\theta_{d/2-1} \cdot c)
    \end{pmatrix},
\]
where
\[
r(\alpha)
=
\begin{pmatrix}
\cos \alpha & \sin \alpha \\
- \sin \alpha & \cos \alpha
\end{pmatrix}
\in \R^{2 \times 2}
\]
is the rotation matrix of angle $\alpha$.

While $\theta$ may be learned, in this work we use the axial RoPE formulation of \citet{Heo2024}, which corresponds to:
\[
\theta = \concat_{i \in [1,p]} \concat_{j \in [0,(d/2-1)/p]} \left( 10000^{-2jp/d} \, e_i \right),
\]

RoPE embeds coordinates directly into the attention logits by separately modifying the query and key vectors:
\[
    q^{\mathrm{RoPE}}(c) = R_\theta(c)q(c),
    \qquad
    k^{\mathrm{RoPE}}(\xi) = R_\theta(\xi)k(\xi).
\]

Due to the properties of rotation matrices, the resulting attention logits depend only on the relative displacement $c-\xi$:
\[
q^{\mathrm{RoPE}}(c)^\top\, k^{\mathrm{RoPE}}(\xi)
=
q^\top(c)\, R^\top_\theta(c-\xi)\, k(\xi).
\]

A direct calculation shows that RoPE corresponds to a Fourier decomposition of the attention logits over the frequency basis $\theta$, with modulated coefficients $\alpha_j(q(c),k(\xi))$ and $\beta_j(q(c),k(\xi))$:
\[
\begin{split}
& \alpha_j(q(c),k(\xi)) = q_{2j}(c) k_{2j}(\xi) + q_{2j+1}(c) k_{2j+1}(\xi), \\
& \beta_j(q(c),k(\xi)) = q_{2j+1}(c) k_{2j}(\xi) - q_{2j}(c) k_{2j+1}(\xi), \\
& z(q(c), k(\xi), c-\xi) = \sum_{j=0}^{d/2-1} \alpha_j(q(c),k(\xi)) \cos\!\left(\theta_j \cdot (c-\xi)\right) + \beta_j(q(c),k(\xi)) \sin\!\left(\theta_j \cdot (c-\xi)\right).
\end{split}
\]

\subsection{Local positional embeddings} \label{app:local-pe}

We build - on top of RoPE and FlashBias \citep{Wu2025} - a localized embedding. We specifically separate the expressive relative attention term (handled by RoPE) and a purely geometrical locality potential $\Phi$ as an additive FlashBias-like term:
\[
z(q(c), k(\xi), \delta) = q^\top(c)R^\top_\theta(\delta) k(\xi) - \sqrt d \Phi(\delta),
\]
where $\delta = c-\xi$ is the relative displacement.

\paragraph{Anisotropic L2 locality potential}

A simple potential function can be written using the anisotropic L2 distance. With $A=\mathrm{diag}(\lambda_1^{-2}, \dots \lambda_p^{-2})$, the resulting potential is defined as:
\[
\Phi^\mathrm{L2}_A(\delta) = \frac{1}{2} \delta^\top A \delta = \frac{1}{2} \sum_{i=1}^p (\delta\cdot e_i)^2 / \lambda_i^2.
\]

Each term of this potential is exactly separable with the following identity:
\[
(c_i-\xi_i)^2 = \begin{pmatrix} c_i^2 & -2c_i & 1 \end{pmatrix} \begin{pmatrix} 1 \\ \xi_i\\ \xi_i^2\end{pmatrix}.
\]

This lets us augment the standard RoPE-modulated embeddings by concatenating $3p$ additional locality channels. Hence we define the query as:
\[
q^{\mathrm{L2}}(c)
=
R_\theta(c)q(c)
\concat_{i=1}^p
\frac{1}{2}  \begin{pmatrix} c_i^2/\lambda_i^2 \\ -2c_i/\lambda_i \\ 1 \end{pmatrix}.
\]
and the key embedding is defined as
\[
k^{\mathrm{L2}}(\xi)
=
R_\theta(\xi)k(\xi)
\concat_{i=1}^p
\begin{pmatrix} 1 \\ \xi_i / \lambda_i\\ \xi_i^2 / \lambda_i^2 \end{pmatrix}.
\]
Both embeddings lie in $\R^{d+3p}$.

\paragraph{Anisotropic and asymmetric hyperbolic locality potential.}

Let's now define a directional and anisotropic locality potential acting on relative displacements:
\[
\Phi^\mathrm{hyp}_{A_+, A_-}(\delta)
=
\frac{1}{2}
\sum_{i=1}^{p}
\left[
\exp\!\left(A_-^{1/2} \delta \cdot e_i\right)
+
\exp\!\left(-A_+^{1/2} \delta \cdot e_i\right)
\right].
\]

This potential is a sum of direction-wise exponential functions encoding anisotropic and asymmetric interaction ranges in physical space through diagonal matrices $A_+=\mathrm{diag}(\lambda_{1,+}^{-2}, \dots \lambda_{p,+}^{-2})$ and $A_-=\mathrm{diag}(\lambda_{1,-}^{-2}, \dots \lambda_{p,-}^{-2})$.

For each spatial direction $i$, the exponential terms are exactly multiplicatively separable:
\[
\exp\!\left(\frac{c_i-\xi_i}{\lambda_{i,-}}\right)
=
\exp\!\left(\frac{c_i}{\lambda_{i,-}}\right)
\exp\!\left(-\frac{\xi_i}{\lambda_{i,-}}\right),
\]
\[
\exp\!\left(-\frac{c_i-\xi_i}{\lambda_{i,+}}\right)
=
\exp\!\left(-\frac{c_i}{\lambda_{i,+}}\right)
\exp\!\left(\frac{\xi_i}{\lambda_{i,+}}\right).
\]

Similarly to the L2 bias, this lets us augment the standard RoPE-modulated embeddings by concatenating $2p$ additional locality channels, corresponding to the positive and negative directions of each spatial axis.

The query embedding is defined as
\[
q^{\mathrm{hyp}}(c)
=
R_\theta(c)q(c)
\concat_{i=1}^p
-\frac{\sqrt d}{2}  \; \exp(- c_i /\lambda_{i,+})
\concat_{i=1}^p -\frac{\sqrt d}{2}
\exp(+ c_i/\lambda_{i,-}),
\]
and the key embedding is defined as
\[
k^{\mathrm{hyp}}(\xi)
=
R_\theta(\xi)k(\xi)
\concat_{i=1}^p
 \exp(+ \xi_i/\lambda_{i,+})
\concat_{i=1}^p
\exp(- \xi_i/\lambda_{i,-}).
\]

Both embeddings lie in $\mathbb{R}^{d+2p}$.

\paragraph{Hyperbolic bias is a natural extension of L2 bias for asymmetric cases}.

Let us consider the behavior of hyperbolic embeddings for a symmetric choice of interaction ranges such that $A_+=A_-=A$.

Then the hyperbolic bias expression simplifies with a $\cosh$ function:
\[
\Phi^\mathrm{hyp}_A = \sum_{i=1}^p \cosh \left( (\delta\cdot e_i/) \lambda_i \right).
\]

Assuming small values of displacements $\delta$, for which $\delta \cdot e_i \ll \lambda_i$ and $\cosh(\delta_i/\lambda_i) \sim 1 + (\delta_i/\lambda_i)^2/2$, one gets,
\[
\Phi^\mathrm{hyp}_A(\delta) \sim  p + \frac{1}{2} \sum_{i=1}^p \frac{\delta_i^2}{\lambda_i^2} \sim p + \Phi_A^\mathrm{L2}(\delta).
\]

Hence, for small displacements, both embeddings are similar up to an additive constant, which disappears under softmax normalization. 

\begin{figure}[h]
    \centering
    \includegraphics[width=0.5\linewidth]{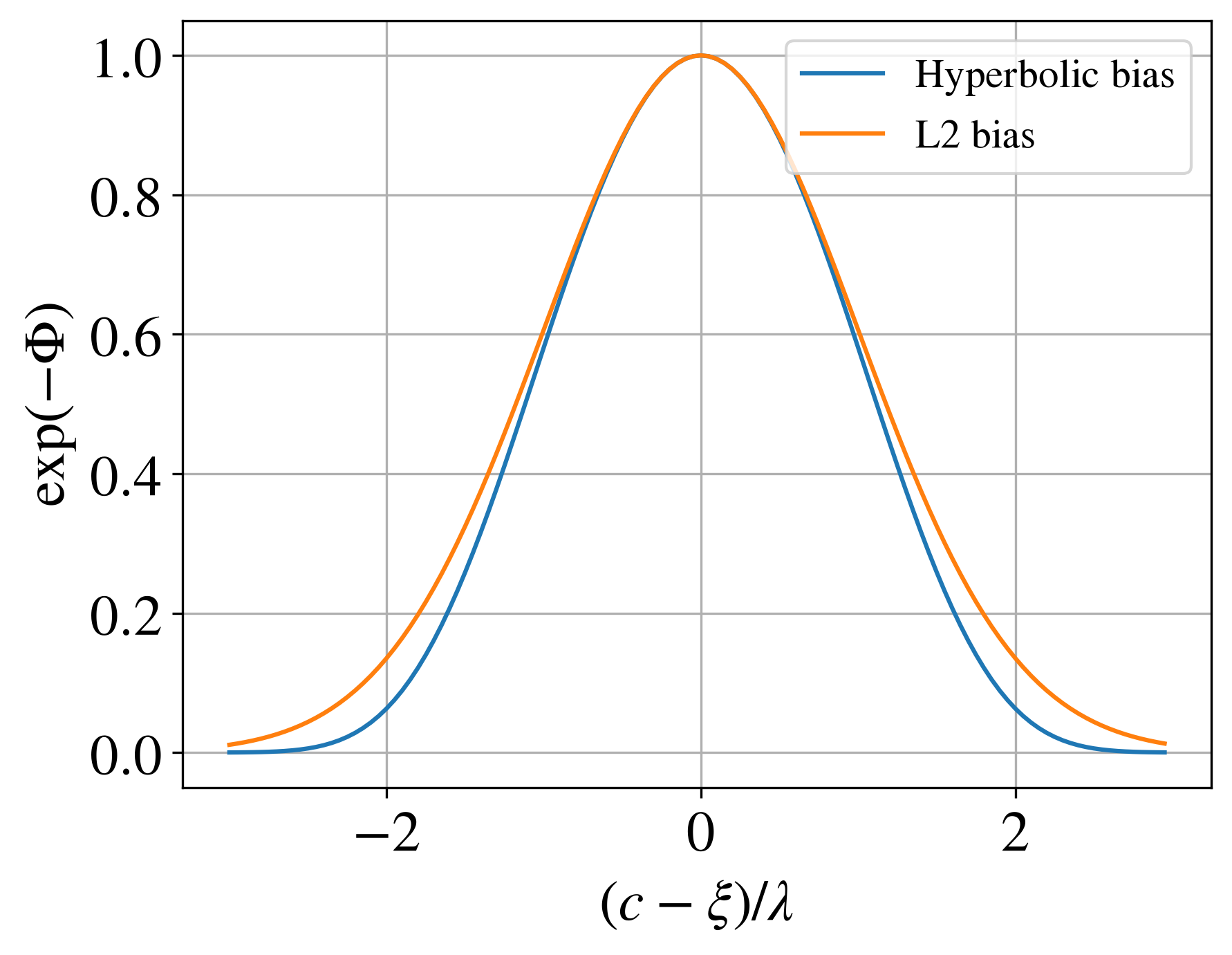}
    \caption{Comparison of the shape of the kernel enforced by an hyperbolic and an L2 bias, not accounting for the RoPE term, and assuming a 1D symmetric case.}
    \label{fig:potentials}
\end{figure}

Furthermore, both potentials will diverge to $+\infty$ for large values of displacements. This leaves only a difference in the transitory region: a kernel with an hyperbolic bias has a slightly faster decay than with an L2 bias (see Figure \ref{fig:potentials}). Consequently, we consider hyperbolic biases to naturally extend L2 biases to asymmetric cases.

% Hyperparameters $\{\lambda_1, \dots \lambda_p\}$ allows a fine control of the potential behavior for each axis. 

%Perhaps too redundant from main paper work?
% Furthermore, softmax normalization results in the attention kernel multiplied by a multivariate gaussian function $f$ with standard deviations $\{\lambda_1, \dots \lambda_p\}$:

% \[
% f_{\{\lambda_i\}}(\delta) = \exp(-\sum_{i=1}^p (\delta\cdot e_i)^2 / \lambda_i^2)
% \]

% \[
% \kappa(q(c), k(\xi), c, \xi) = \frac{f_{\{\lambda_i\}}(\delta) e^{q(c)R_\theta(c-\xi)k(\xi)}}{\int_\Omega  f_{\{\lambda_i\}}(\delta)e^{q(c)R_\theta(c-\xi)k(\xi)}}
% \]

% It is thus natural to interpret $\{\lambda_i\}$ as standard deviations controlling the kernel support.

\section{Numerical stability} \label{app:num-stability}

\begin{figure}[h]
    \centering
    \includegraphics[width=\linewidth]{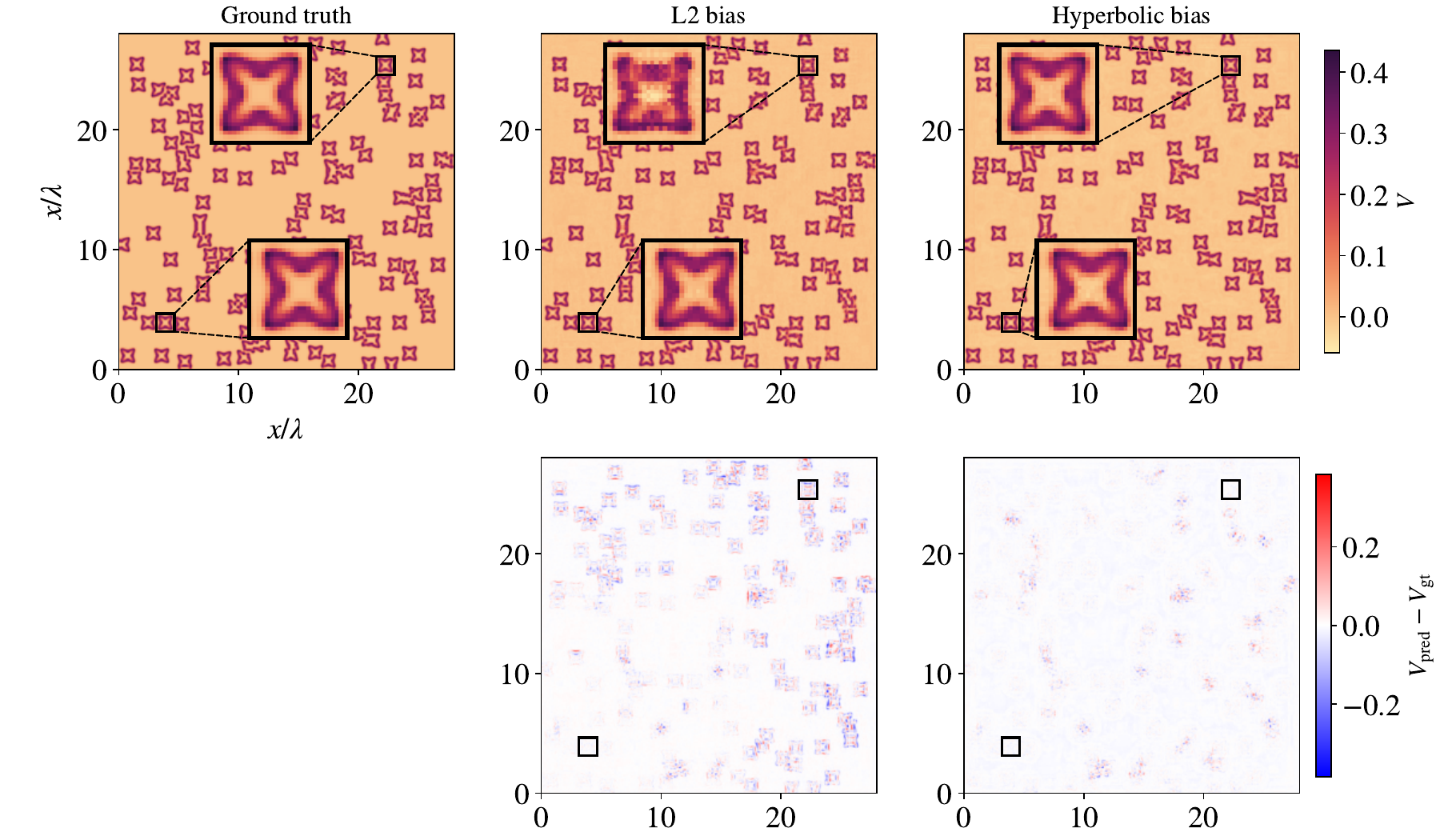}
    \caption{Comparative predictions of hyperbolic and L2 biases at $t=1$ on the Gray Scott dataset for a scaling $s_x \times s_y = 7 \times 7$ with BFloat16 precision. L2 biases exhibits strong numerical errors for larger coordinates values, while hyperbolic biases remains stable.}
    \label{fig:num-err}
\end{figure}

\paragraph{Numerical errors of L2 decomposition}
We found experimentally that an L2 bias tends to exhibit numerical instabilities when reaching high coordinates values. Figure \ref{fig:num-err} shows this numerical error on the Gray Scott dataset for a scaling $s_x \times s_y = 7 \times 7$. The error of the L2 model's prediction is non-homogeneous and increases with the distance to the origin. By contrast, an hyperbolic bias remains stable even for large coordinate values.

\begin{figure}[h]
    \centering
    \includegraphics[width=\linewidth]{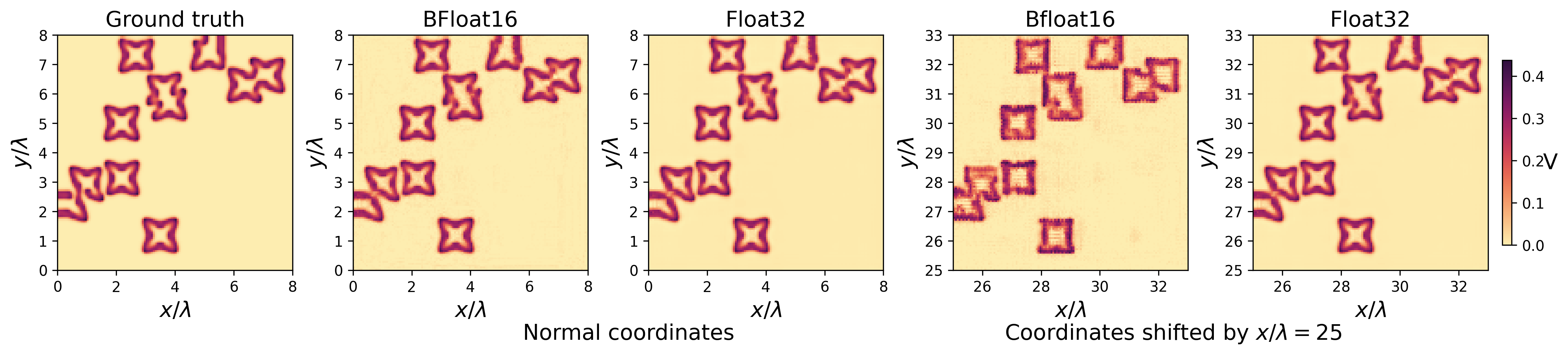}
    \caption{Numerical stability experiment. Left: Predictions using an L2 bias for Bfloat16 and Float32 precisions for coordinate values near the origin. Right: Predictions for coordinate values shifted by $x/\lambda=25$ and $y/\lambda=25$, mimicking values seen at large scaling.}
    \label{fig:num-err-2}
\end{figure}

To demonstrate that this error indeed comes from the numerical precision, we compare predictions using Bfloat16 and Float32 formats. A scale $7$ inference is not computationally feasible with Float32. We instead perform this experiment on $s_x \times s_y = 2 \times 2$ and shift the input coordinates to reproduce coordinate values seen at larger scales. We present the result in Figure \ref{fig:num-err-2}. For small coordinates, both predictions yield similar results. However, for large coordinates, prediction accuracy degrades significantly when using BFloat16 precision, while remaining stable with Float32.

\paragraph{Numerical overflow limit of hyperbolic biases} The presented asymmetric bias relies on exponential functions, which introduces numerical limitations due to potential overflow and underflow. Specifically, the computation of the bias terms will overflow or underflow on positions that are too far from the domain center. Let $v_{\min}$ and $v_{\max}$ denote the smallest and largest representable positive numbers for a given precision. The maximum attainable domain size scales as
\[
\lambda \bigl(\log v_{\max} - \log v_{\min}\bigr).
\]
Table~\ref{tab:scaling-limits} summarizes the resulting limits for common floating-point formats, assuming a $1D$ isotropic and symmetric bias controlled by a decay length $\lambda$. Commonly used Bfloat16 format presents a maximum scaling of $181 \lambda$, which is very substantial in the context of physics simulations and covers most use cases.

\begin{table}[h] %TODO: check computations
    \centering
    \caption{Maximum attainable scaling range of hyperbolic biases before numerical overflow or underflow.}
    \label{tab:scaling-limits}
    \begin{tabular}{cccc}
        \toprule
        Precision & underflow limit & overflow limit & max. domain size \\
        \midrule
        Float16 & $6\times10^{-8}$ & $6.5\times10^{4}$ & $27\,\lambda$ \\
        Float32 / Bfloat16 & $6\times10^{-41}$ & $3\times10^{38}$ & $181\,\lambda$ \\
        Float64 & $2\times10^{-308}$ & $2\times10^{308}$ & $1418\,\lambda$ \\
        \bottomrule
    \end{tabular}
\end{table}

\section{Training setup and model hyperparameters} \label{app:training}

This appendix presents detailed training setup and model hyperparameters. All conducted experiments were run using the Noether framework \citep{Bleeker2026}.

% We use the same training setup for all experiments, only varying the number of training epochs between academic cases and AB-SWIFT. We give implementation details below and report hyperparameters in Table \ref{tab:train-params}.

For all experiments, training is performed using the Lion optimizer \citep{Chen2023a}, with a $5\%$ linear learning‑rate warmup up to $5\times10^{-5}$, followed by a cosine decay to $1\times10^{-6}$. The weight decay parameter of Lion is set to $0.05$. All models are trained using Bfloat16 precision. We train for $100$ epochs with a batch size of $32$ for the academic benchmarks, and for $500$ epochs with a batch size of $1$ for AB‑SWIFT.

Input and output features are standardized, and positions are normalized anisotropically to $[0,1000]$ in the training dataset. A log scale is used on the turbulence-related variables $k$ and $\epsilon$ predicted by AB-SWIFT, following the choices of \citet{Villeroche2026}.

\paragraph{Academic case models} We apply the same model for each dataset, only varying the positional embedding between experiments. We build a generic transformer neural operator architecture consisting of stacked transformer blocks and linear input embeddings and output projections. Each transformer block consists of a self-attention layer followed by a feedforward layer, with residual connections between each sublayer. We report the associated hyperparameters in Table \ref{tab:aca-params}.

\begin{table}[h]
    \centering
    \caption{Hyperparameters of the models used for all academic datasets experiments.}
    \label{tab:aca-params}
    \begin{tabular}{lr} \toprule
         Parameter & value \\ \midrule
         Hidden dimension & 192 \\
         Number of transformer blocks & 6 \\
         Number of attention heads & 3 \\
         Feedforward hidden layers & 1 \\
         Feedforward expansion factor & 4 \\
         Feedforward activation & GeLU \\
         RoPE max frequency & 10000 \\
         \bottomrule
    \end{tabular}
\end{table}

\paragraph{AB-SWIFT} We re-implement AB-SWIFT using the Noether framework, following the original implementation of \citet{Villeroche2026}. However, unlike the original implementation, we do not use absolute positional embeddings to initially encode the latent state from absolute positions, and instead rely only on geometric information when available, or generate constant initial embeddings when no geometric information are available. We report details on hyperparameters in Table \ref{tab:abswift-params}.

\begin{table}[h]
    \centering
    \caption{AB-SWIFT hyperparameters.}
    \label{tab:abswift-params}
    \begin{tabular}{lr} \toprule
         Description & Value \\ \midrule
        \multicolumn{2}{c}{Model parameters} \\ \addlinespace
        Hidden dimension & 192 \\
        Obstacles supernode pooling radius & 1 \\
        Terrain supernode pooling radius & 5 \\
        Number of processor blocks & 3 \\
        Number of decoder blocks & 4 \\
        Number of attention head & 3 \\
        Feedforward hidden layers & 1 \\
        Feedforward expansion factor & 4 \\
        Feedforward activation & GeLU \\
         RoPE max frequency & 10000 \\
        \midrule 
        \multicolumn{2}{c}{Training pipeline parameters} \\ \addlinespace
        Number of points describing obstacles & 4096 \\
        Number of points describing the terrain & 4096 \\
        Number of obstacles supernodes & 1024 \\
        Number of terrain supernodes   & 1024 \\
        Number of volume anchor point & 8192 \\
        \midrule
        \multicolumn{2}{c}{Inference pipeline parameters} \\ \addlinespace
         \multicolumn{2}{c}{All number of points are scaled proportionally to the area of the inference domain.} \\
        \bottomrule
    \end{tabular}
\end{table}

\paragraph{Training cost}
Training takes approximately $\qty{1}{h}$ on a A100 GPU for the shallow water dataset, and $\qty{1.5}{h}$ for the Gray-Scott and the microscale atmospheric flow dataset. Running all experiments sequentially takes approximately $\qty{13}{h}$.

\section{Datasets} \label{app:datasets}

This appendix further describes the datasets used in this paper, summarized in Table \ref{tab:datasets}. 

\begin{table}[h]
    \caption{Datasets considered in this work. For each dataset, we generate a large number of small-scale simulations for training, and a small number of large-scale simulations for evaluations, with a per-axis scaling ranging from $1$ to $10$. $\Delta\cdot$ denote the difference between two time steps.}
    \label{tab:datasets}
    \centering
    \small
    \begin{tabular}{cccm{3.1cm}m{2.3cm}} \toprule
         Dataset & PDE & Dim & input variable & predicted variable \\ \midrule
         SWE 1D & Shallow water & 1D+time & Water height $h$ and velocity $v$ & $\Delta h, \Delta v$ \\ 
         GrayScott & Gray Scott & 2D+time & Concentrations $U$ and $V$ & $\Delta U, \Delta V$ \\  \\
         RandomBuildingsDataset & Navier-Stokes & 3D & Building topography \newline meteorological conditions & 3D steady-state atmospheric flow variables \\ % \newline $\mathbf{v},p,\theta,k,\epsilon$ \\
         \bottomrule
    \end{tabular}
\end{table}

\subsection{Shallow water equations}

Shallow water equations represent the movements of waves on a water surface, assuming a low water depth relative to the surface. The shallow water equations solved reads:
\[
    \frac{\partial h}{\partial t} + \frac{\partial(hv)}{\partial x} = 0,
    \qquad
    \frac{\partial(hv)}{\partial t} + \frac{\partial(hv^2)}{\partial x} + gh\frac{\partial h}{\partial x} = 0,
\]
where $h$ and $v$ respectively represent the water height and velocity, and $g=\qty{9.81}{\m\per\s\squared}$ is the gravity acceleration constant. Homogeneous Neumann boundary conditions on $h$ and $v$ are used.

We generate an initial dataset of $\num{800}$ training, $\num{100}$ test and $\num{100}$ validation simulations of a domain of width $\qty{100}{\m}$, and an initial water height $h_0 = \qty{1}{\m}$. We randomly place $3$ crenel perturbations of the initial water height in each simulation, with widths and heights uniformly sampled respectively between $\qtyrange{4}{15}{\m}$ and  $\qtyrange{0.02}{0.08}{\m}$. We simulate the obtained system for $\qty{15}{\s}$, and save the state every $\qty{0.3}{\s}$.

We then generate large scale splits for scalings $s=2,3,7$ and $10$, and run $50$ simulations per split. We scale the domain width and the number of initial perturbations by $s$.

Data generation takes a few seconds on an Intel i7-11850H laptop CPU.

\subsection{Gray-Scott equations}

The Gray-Scott equation system describes the evolution of two chemical substances $U$ and $V$, with $U$ being consumed by $V$ and both substances diffusing over time. The Gray-Scott equation systems reads:
\[
    \frac{\partial U}{\partial t} = D_u \nabla^2 U - U V^2 + F (1 - U),
    \qquad
    \frac{\partial V}{\partial t} = D_v \nabla^2 V + U V^2 - (F + k) V,
\]
where $D_u=0.2$ and $D_v=0.1$ are respectively the diffusion coefficients of $U$ and $V$, $k=0.06$ is the kill rate of $V$ and $F=0.035$ is the feed rate. We use homogeneous Neumann boundary conditions on $U$ and $V$.

We generate an initial dataset of $\num{800}$ training $\num{100}$ test and $\num{100}$ validation simulations of a square domain of width $128 \times 128$ width grid cells of size $2\times2$ and, with homogeneous initial concentrations $U=1$ and $V=0$. For each simulation, we randomly place $3$ initial sources of $V$ within the simulated domain on squares of width $5$ where we set $U=0.5$ and $V=0.25$. We simulated the obtained system for $\num{5000}$ steps and save the state every $\num{500}$ steps.

We then generate large scale splits for scalings $s=2,3,7$ and $10$, and run respectively $50$, $50$, $25$, $10$ simulations per scaling. We scale the domain width and height by $s$ and the number of initial perturbations by $s^2$.

Data generation takes around $\qty{20}{\min}$ on an Intel i7-11850H laptop CPU.

\subsection{Random buildings dataset}

The random buildings datasets contain steady-state atmospheric flows around randomly generated urban geometries with varied building geometries. Readers are referred to the original work of \citet{Villeroche2026} for the detailed CFD simulation setup.

We generate large-scale validation cases by scaling the size of the buildings area width by $s$. For each simulation, we generate a new urban geometry on this enlarged area using the same generation method as \citet{Villeroche2026}. We scale the number of buildings by $s^2$, keeping the building density constant across all scalings. We generate large scale splits for scalings $s=2$ and $5$ and generate $10$ simulations per scaling.

We restrict ourselves to neutral atmospheric stratifications because buildings are placed very close to the upstream  boundary in the original dataset, making boundary conditions interact strongly with the buildings in other stratifications. As this effect depends on the exact distance between the buildings and the upwind boundary of the simulation domain, it could not be robustly reproduced when increasing the size of the mesh. In neutral stratification, this effect is much less pronounced. We note that this is a limit of the CFD setup and not of the proposed machine learning approach.

Data generation of large scale cases takes around $\qty{11}{h}$ on $4$ Intel Platinum 8260 CPUs.

\section{Visualization of datasets and predictions} \label{app:visu}

The appendix presents visualizations of the different datasets and predictions.

\subsection{Shallow water equations}

\begin{figure}[h]
    \centering
    \includegraphics[width=0.8\linewidth]{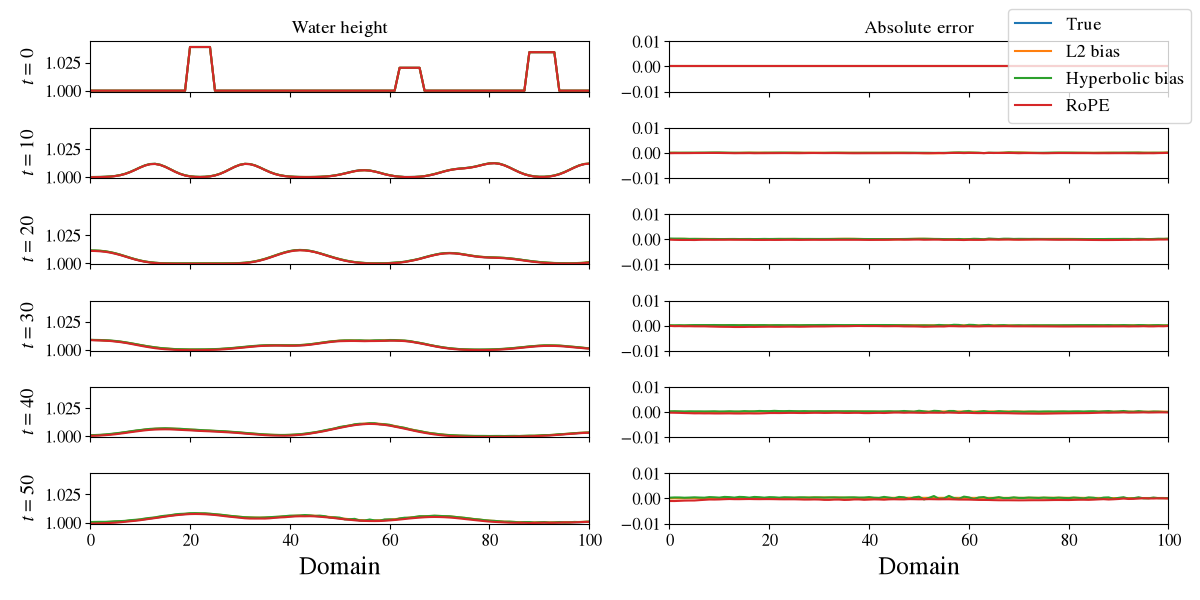}
    \caption{Shallow water ground truth and predictions for $s=1$.}
    \label{fig:swe-visu-1}
\end{figure}

\begin{figure}[h]
    \centering
    \includegraphics[width=0.8\linewidth]{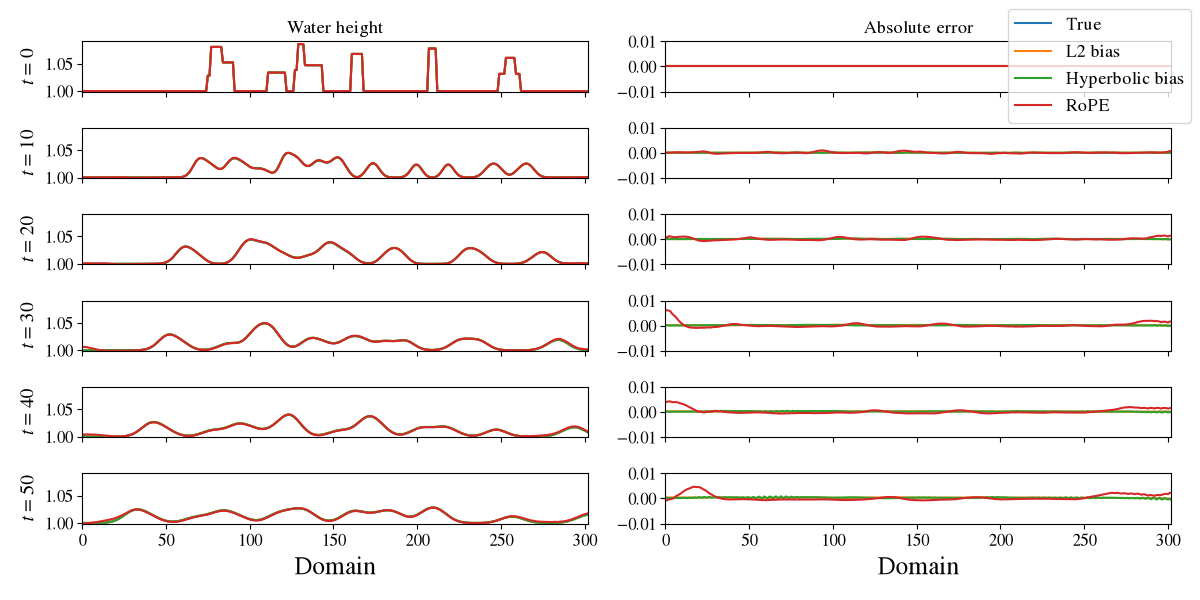}
    \caption{Shallow water ground truth and predictions for $s=3$.}
    \label{fig:swe-visu-3}
\end{figure}

\begin{figure}[h]
    \centering
    \includegraphics[width=0.8\linewidth]{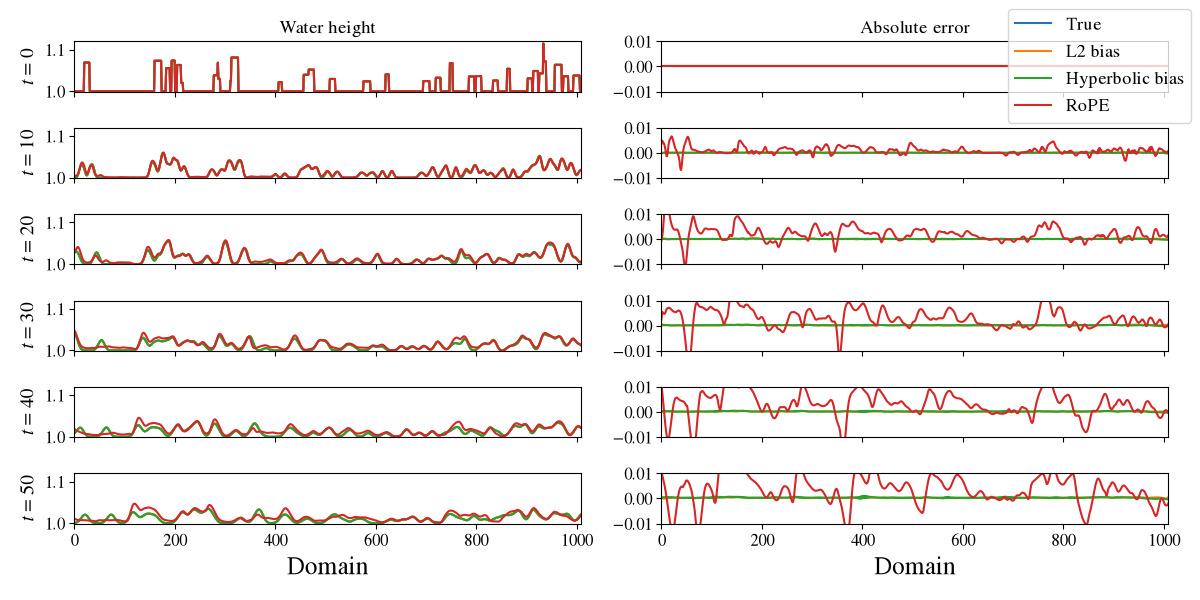}
    \caption{Shallow water ground truth and predictions for $s=10$.}
    \label{fig:swe-visu-10}
\end{figure}

Figures \ref{fig:swe-visu-1},  \ref{fig:swe-visu-3}, and \ref{fig:swe-visu-10} present ground truth and models predictions for scalings $s=1,3$ and $10$ respectively.

\subsection{Gray-Scott equations}

\begin{figure}[h]
    \centering
    \includegraphics[width=0.8\linewidth]{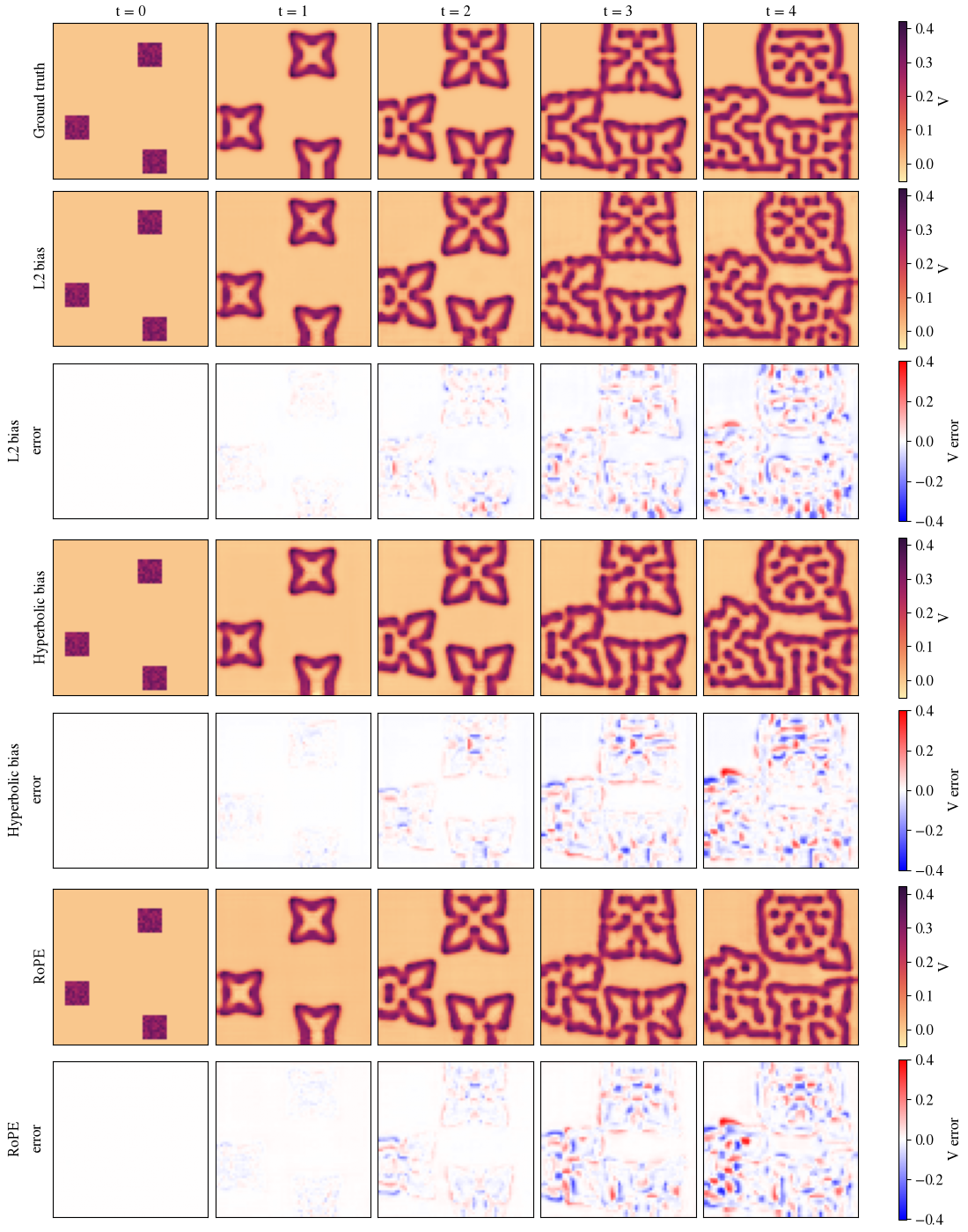}
    \caption{Gray-Scott ground truth and predictions for $s_x \times s_y = 1 \times 1$.}
    \label{fig:gs-visu-1}
\end{figure}

\begin{figure}[h]
    \centering
    \includegraphics[width=0.8\linewidth]{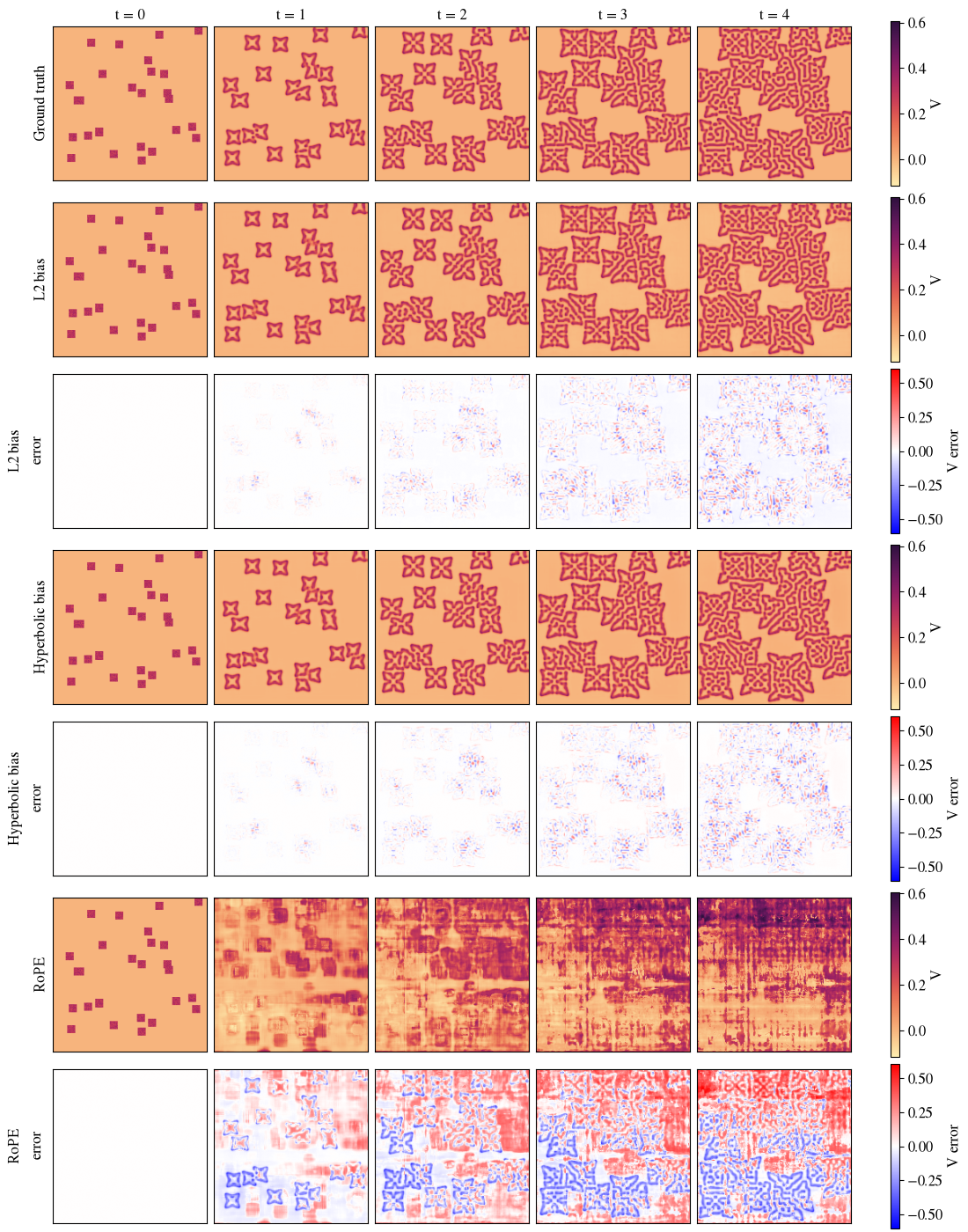}
    \caption{Gray-Scott ground truth and predictions for $s_x \times s_y = 3 \times 3$.}
    \label{fig:gs-visu-3}
\end{figure}

\begin{figure}[h]
    \centering
    \includegraphics[width=0.8\linewidth]{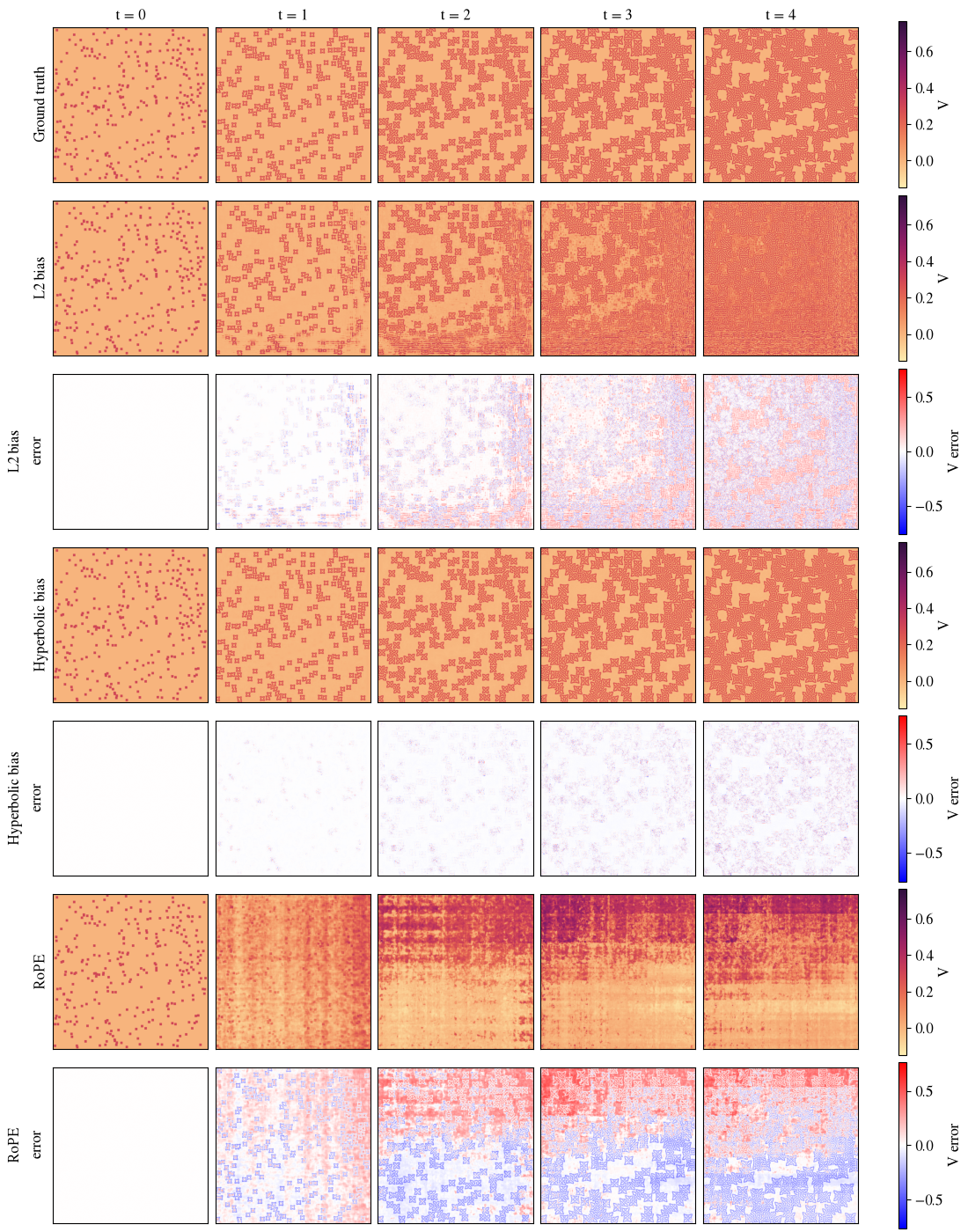}
    \caption{Gray-Scott ground truth and predictions for $s_x \times s_y = 10 \times 10$.}
    \label{fig:gs-visu-10}
\end{figure}

Figures \ref{fig:gs-visu-1},  \ref{fig:gs-visu-3}, and \ref{fig:gs-visu-10} present ground truth and models predictions for scalings $s_x \times s_y=1 \times 1,3 \times 3$ and $10 \times 10$ respectively.

\subsection{Microscale atmospheric flow}

\begin{figure}[h]
    \centering
    \includegraphics[width=0.8\linewidth]{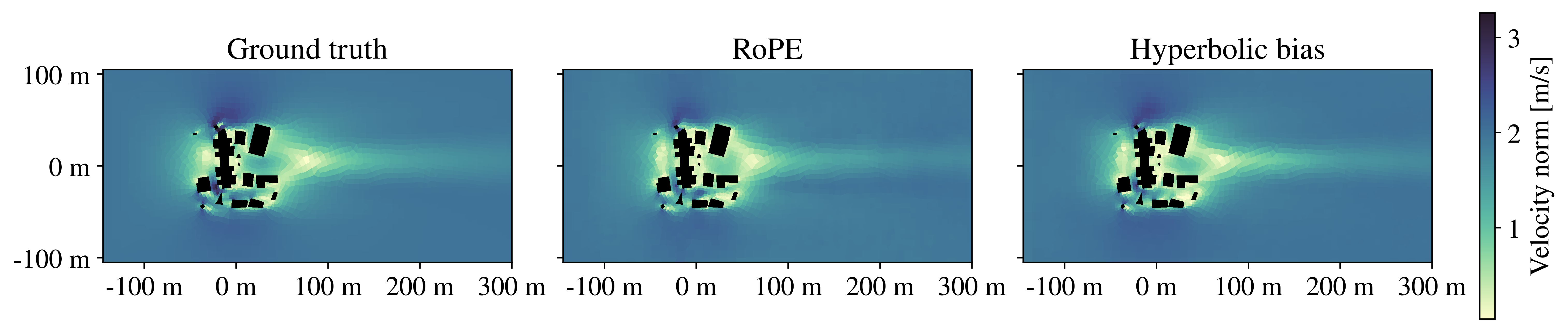}
    \caption{Microscale atmospheric flow ground truth and prediction for $s_x \times s_y = 1 \times 1$.}
    \label{fig:abswift-visu-1}
\end{figure}

\begin{figure}[h]
    \centering
    \includegraphics[width=0.8\linewidth]{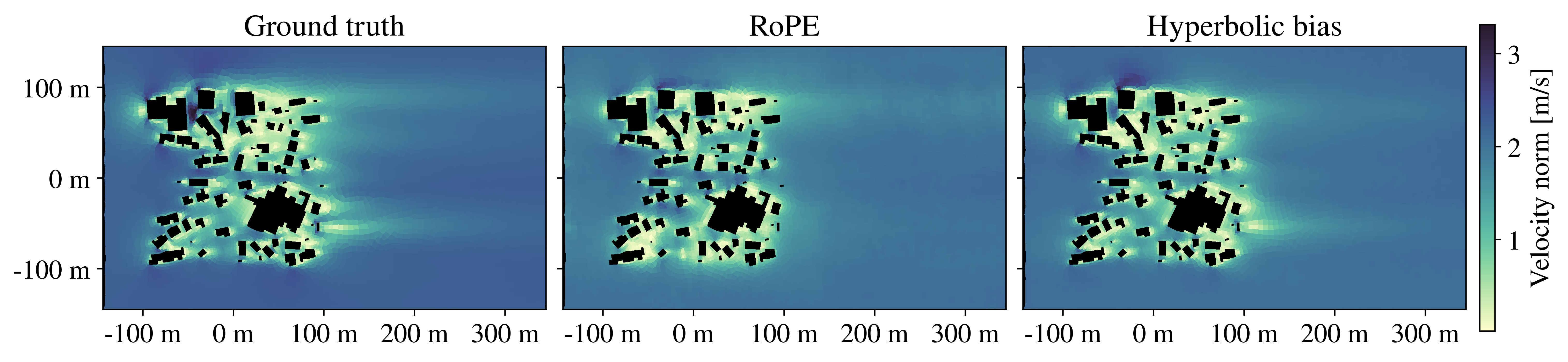}
    \caption{Microscale atmospheric flow ground truth and prediction for $s_x \times s_y = 2 \times 2$.}
    \label{fig:abswift-visu-2}
\end{figure}

\begin{figure}[h]
    \centering
    \includegraphics[width=0.8\linewidth]{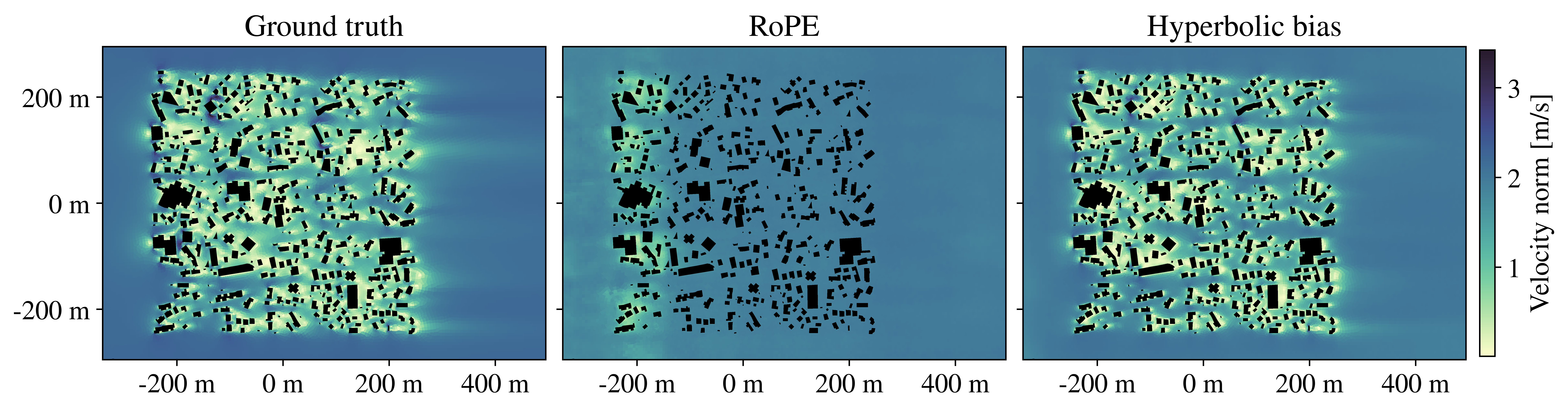}
    \caption{Microscale atmospheric flow ground truth and prediction for $s_x \times s_y = 5 \times 5$.}
    \label{fig:abswift-visu-5}
\end{figure}

Figures \ref{fig:abswift-visu-1},  \ref{fig:abswift-visu-2}, and \ref{fig:abswift-visu-5} presents ground truth and models predictions for scalings $s_x \times s_y=1 \times 1,2 \times 2$ and $5 \times 5$ respectively.

\end{document}